\documentclass[sigconf]{acmart}
\usepackage{listings}
\usepackage{booktabs}
\usepackage{array}
\usepackage{multirow}
\usepackage{enumitem}
\usepackage[linesnumbered,ruled,vlined]{algorithm2e}
\usepackage{lineno}
\usepackage[toc,page]{appendix}
\DeclareUnicodeCharacter{2606}{\ensuremath{\star}}
\usepackage[utf8]{inputenc}
\usepackage[T1]{fontenc}
\PassOptionsToPackage{table}{xcolor}
\usepackage{tcolorbox}

\renewcommand\footnotetextcopyrightpermission[1]{} 

\AtBeginDocument{%
  \providecommand\BibTeX{{%
    \normalfont B\kern-0.5em{\scshape i\kern-0.25em b}\kern-0.8em\TeX}}}

\graphicspath{{./images/}} 
\begin{document}

\title{Exploring a Hybrid Deep Learning Approach for Anomaly Detection in Mental Healthcare Provider Billing: Addressing Label Scarcity through Semi-Supervised Anomaly Detection}

\author{
  Samirah Bakker$^{1}$, 
  Yao Ma$^{1}$*, 
  Seyed Sahand Mohammadi Ziabari$^{1,2}$
}

\affiliation{%
  \institution{
  $^{1}$Informatics Institute, University of Amsterdam, 1098XH Science Park, Amsterdam, The Netherlands}
  \country{}
}

\affiliation{%
  \institution{$^{2}$Department of Computer Science and Technology, SUNY Empire State University, Saratoga Springs, NY, USA}
   \country{}
}

\email{samirah.bakker@student.uva.nl, y.ma3@uva.nl,sahand.ziabari@sunyempire.edu}

\begin{abstract}

The complexity of mental healthcare billing enables anomalies, including fraud. While machine learning methods have been applied to anomaly detection, they often struggle with class imbalance, label scarcity, and complex sequential patterns. This study explores a hybrid deep learning approach combining Long Short-Term Memory (LSTM) networks and Transformers, with pseudo-labeling via Isolation Forests (iForest) and Autoencoders (AE). Prior work has not evaluated such hybrid models trained on pseudo-labeled data in the context of healthcare billing. The approach is evaluated on two real-world billing datasets related to mental healthcare. The iForest LSTM baseline achieves the highest recall (0.963) on declaration-level data. On the operation-level data, the hybrid iForest-based model achieves the highest recall (0.744), though at the cost of lower precision. These findings highlight the potential of combining pseudo-labeling with hybrid deep learning in complex, imbalanced anomaly detection settings.

\end{abstract}



\pagestyle{plain}
\setcounter{page}{0}

\maketitle

\section{Introduction}
\label{sec:introduction}

The complexity of healthcare billing and large insurance volumes create an environment where anomalies, including fraud, can thrive \cite{hamid-2024}. Fraud by healthcare providers results in significant financial losses and negatively impacts healthcare systems \cite{najar-2025}. In the Netherlands, healthcare fraud is estimated to cost billions of euros annually \cite{Rekenkamer2022}, burdening insurers. Despite ongoing efforts, the likelihood of identifying perpetrators remains low, underscoring the need for more effective detection methods.

Mental healthcare billing, in particular, has proven vulnerable to fraud due to the complexity of diagnoses and treatments, as well as the difficulty of verifying the provided services. Investigations have exposed fraudulent practices, including billing for unprovided claims, inflated rates, and double-charging for services \cite{riecon, smolenaars-2024}. These cases highlight the urgent need for fraud detection methods to prevent financial abuse.

A fundamental challenge in fraud detection is the highly imbalanced data, where normal transactions significantly outnumber fraudulent cases. This imbalance often leads to performance bias \cite{cao2019learning}, especially since labeled fraud cases are typically scarce in practical applications \cite{motie-2023}. In this context, fraud can be framed as an anomaly detection problem, where the goal is to find rare patterns in the data that deviate from the expected behavior. To address these challenges, this study adopts a pseudo-labeling strategy based on unsupervised anomaly detection. Instead of relying on a small number of labeled fraud cases, unsupervised models identify structural deviations in the data to generate synthetic labels, thereby enriching the minority class. Recent studies have shown that such strategies can enhance classification performance across multiple domains where labeled data is scarce \cite{li-2023, han-2023, dubourvieux-2022}. 

Although widely used, machine learning methods face increasing challenges as fraudulent techniques evolve and high-dimensional data grows. As a result, these methods struggle to identify these anomalies over time \cite{tang-2024}. By contrast, deep learning models have consistently outperformed conventional machine learning algorithms in fraud detection tasks due to their ability to learn complex, high-dimensional relationships \cite{ashtiani-2022, kim-2019}. In particular, Long Short-Term Memory (LSTM) networks are well-suited for detecting temporal anomalies in sequential data \cite{mienye-2024}. In addition, Transformers, due to their self-attention mechanism, can effectively capture complex dependencies and interactions within the data \cite{yu-2024}. Their robust performance across domains has been demonstrated in various anomaly detection tasks. Combining these two approaches can offer an even more effective solution, as it builds on the strengths of each model. While LSTMs and Transformers have been applied separately for anomaly detection, their combined use remains largely unexplored in the context of imbalanced mental healthcare billing data. Moreover, the application of pseudo-labeling with hybrid models in this domain is understudied.

This study proposes a hybrid deep learning framework that combines LSTM networks and Transformer architectures to improve anomaly detection in imbalanced mental healthcare billing data. The approach addresses key challenges in this domain, including data imbalance, limited labeled instances of anomalous behavior, and the complex temporal and contextual patterns inherent in billing sequences.

To mitigate the issue of label scarcity, the methodology incorporates pseudo-labeling strategies using unsupervised anomaly detection techniques such as Isolation Forests and Autoencoders. These methods generate synthetic anomaly labels to augment the training data, allowing the hybrid model to learn from enriched and more representative datasets. The integration of LSTM and Transformer components enables the model to capture both temporal dependencies and intricate feature interactions, making it particularly suitable for complex anomaly detection tasks.

The proposed framework is evaluated on two real-world datasets drawn from mental healthcare provider billing systems. Experimental results demonstrate that pseudo-labeling significantly enhances model performance across several metrics. In particular, the hybrid model outperforms standalone LSTM or Transformer architectures in detecting irregularities in both declaration- and operation-level data, achieving higher recall in scenarios where precision trade-offs are acceptable.

This work contributes to the literature by demonstrating the viability of combining pseudo-labeling with hybrid deep learning architectures for anomaly detection in complex, high-dimensional, and imbalanced healthcare billing environments. The findings underscore the importance of architecture design and data enrichment strategies in building robust anomaly detection systems where labeled data is limited or incomplete.

The remainder of this study is organized as follows: Section \ref{sec:related_work} presents a review of related work, while Section \ref{sec:methodology} details the methodology. Section \ref{sec:results} presents the results followed by a discussion on the findings and implications in Section \ref{sec:discussion}. Finally, Section \ref{sec:conclusion} provides concluding remarks.

\section{Related Work}
\label{sec:related_work}

Fraud detection is a well-studied application of anomaly detection, where the goal is to correctly classify rare instances that deviate from expected patterns. Despite ongoing advancements, multiple challenges remain, such as annotation delays and constantly evolving fraud patterns \cite{pozzolo-2018}. Additionally, the highly imbalanced nature of fraud datasets, with fraudulent transactions comprising only about 0.2\% of total transactions, remains a significant challenge \cite{kulatilleke-2022}. This section reviews existing machine learning and deep learning approaches for anomaly detection, discusses recent advances in hybrid models, and introduces pseudo-labeling for addressing class imbalance. In doing so, it highlights the research gaps this study aims to address.

\subsection{Machine learning models}

Many machine learning methods, such as supervised learning, learn patterns from historical data. However, they often struggle with high-dimensional and sequential data, real-time fraud detection, and imbalanced datasets, where anomalies affect a small percentage of the data \cite{nasir-2021,craja-2020}. For instance, Jain et al. \cite{Jain2019} found that Artificial Neural Networks (ANN) achieved 99.71\% accuracy in credit card fraud detection. However, they noted that most fraud detection techniques struggle with real-time detection due to the rarity of fraud, often identifying it post-occurrence. Similarly, Sailusha et al. \cite{9121114} achieved 99.91\% accuracy with Random Forest and AdaBoost but faced high false negative rates, advocating for deep learning to improve detection and reduce misclassification. Raghavan and Gayar  \cite{Raghavan2019} showed that Support Vector Machines (SVM) performed best on larger datasets, while Convolutional Neural Networks (CNN) outperformed other deep learning models. Nevertheless, the lack of suitable training data remained a major challenge. Although these methods have shown some success, they struggle to capture the complex and evolving patterns of anomalies in sequential data. These limitations underscore the need for more advanced and flexible approaches to fraud detection.

\subsection{Deep learning models}

A promising area for improving anomaly detection is the use of deep learning models. These models excel at processing high-dimensional data and capturing non-linear relationships, making them well-equipped to identify subtle and hidden patterns that traditional methods often miss \cite{motie-2023}. For more complex tasks like fraud detection, deep learning algorithms generally outperform conventional machine learning algorithms \cite{ashtiani-2022}.

\subsubsection{Recurrent Neural Networks}

Among deep learning models, Recurrent Neural Networks (RNN), such as LSTM models, have shown potential in modeling sequential data \cite{yang-2020, shen-2020}.  Their ability to process high-dimensional, time-dependent data provides an advantage for anomaly detection in complex, real-time environments. RNNs have especially proven effective at handling the complexities of financial transaction data, as demonstrated by Dutta and Bandyopadhyay \cite{dutta-2020}. They proposed a stacked RNN model to monitor and detect fraudulent financial transactions in mobile money systems, which achieved an accuracy of 99.87\%. Similarly, Rosaline et al. \cite{roseline-2022} used an LSTM-RNN for credit card fraud detection, incorporating an attention mechanism to enhance performance. Their results demonstrated the effectiveness of LSTM-RNN in handling complex and interrelated data, outperforming classifiers like Naive Bayes, SVM, and ANN. Benchaji et al. \cite{benchaji-2021} further highlighted the potential of LSTM models for fraud detection by integrating an attention mechanism as well as uniform manifold approximation and projection for feature selection. Their experimental results demonstrated that their model outperformed traditional methods.

Despite improvements such as attention mechanisms, RNNs still face challenges when it comes to understanding long-range dependencies and processing large-scale datasets efficiently \cite{mienye-2024}. In addition, traditional RNNs suffer from the vanishing gradient problem, which makes it difficult for these models to retain information over long sequences. This leads to a loss of context and reduced~performance \cite{tang-yuxuan-2024}. 

\subsubsection{Transformers}

Transformers overcome these challenges by processing entire sequences simultaneously. With their self-attention mechanism, Transformers can capture correlations by assigning different weights to different sections of the input sequence. Their robust performance across various tasks has been successfully applied to multiple domains, including anomaly detection. Transformers have shown great promise, as seen in the work of Tang et al. \cite{tang-2024}, who applied a Transformer-based model for financial fraud detection, outperforming traditional machine learning methods. Taneja et al. \cite{taneja-2025} also proposed a Transformer-based framework using BERT for online recruitment fraud detection. Their model demonstrated superiority over conventional methods, achieving an accuracy score of 99\%. Yu et al. \cite{yu-2024} further demonstrated the effectiveness of Transformer models in credit card fraud detection, highlighting their advanced capability to improve accuracy and robustness. Their experimental results showed that the Transformer outperformed traditional methods, showcasing their potential in anomaly detection.

Nevertheless, a known limitation of Transformers is their quadratic memory complexity within the self-attention mechanism, which can lead to scalability issues. This can be problematic in anomaly detection, where large sequential datasets need to be processed. While both RNNs and Transformers have individual strengths, solely relying on a single model is not sufficient to fully capture the complexities of fraudulent activities. 

\subsubsection{Hybrid models}

To mitigate the limitations of individual architectures, recent studies have explored the effectiveness of hybrid approaches that combine RNNs with Transformers. For instance, Zhou et al. \cite{zhou-2023} demonstrated the potential of these models for multiscanning-based hyperspectral image classification, achieving competitive classification performance. Furthermore, Riaz et al. \cite{riaz-2024} applied a hybrid LSTM-Transformer model for fine-grained suggestion mining, showing its ability to enhance the quality and relevance of mined suggestions. In the context of anomaly detection, Ileberi and Sun \cite{ileberi-2024} successfully combined LSTM, CNN, and Transformers to enhance credit card fraud detection. Their hybrid approach outperformed individual-based learners and traditional methods. By leveraging the strength of each model, they were able to capture both spatial and temporal patterns as well as complex relationships within the data. 

However, to the best of the authors' knowledge, the combination of RNNs and Transformer models for anomaly detection has not yet been explored. Building on previous work, this study proposes a more focused hybrid approach, combining only LSTM and Transformer models. By simplifying the ensemble to only two models, this research aims to reduce complexity while still leveraging the complementary strengths of LSTMs for sequential data processing and Transformers for capturing intricate feature interactions in transaction data.

 \subsection{Pseudo-labeling}

Pseudo-labeling has emerged as a practical approach to mitigate class imbalance in anomaly detection. A common strategy is to apply semi-supervised learning, where a small set of labeled anomalies is used to create pseudo-labels. However, this approach risks producing biased models since it heavily relies on a small amount of labeled data when there is a distribution mismatch between the labeled and unlabeled data \cite{yoon-2022}. Unsupervised strategies address this issue by identifying outliers based on structural deviations present in the data itself, rather than relying on labeled data. 

Prior work has demonstrated that unsupervised pseudo-labeling methods can enhance anomaly detection performance in highly imbalanced datasets. For instance, Al-Lahham et al. \cite{al-lahham-2023} introduced a coarse-to-fine pseudo-labeling framework that identifies anomalous video segments based on the statistical properties of the data. Similarly, Im et al. \cite{im-2024} proposed an unsupervised method that refines pseudo-labels by only using internal data consistency, showing strong results in highly imbalanced settings. Unlike techniques such as SMOTE or random oversampling, which assume a well-defined distribution of the minority class and may introduce noise, pseudo-labeling leverages the inherent structure of the data. This makes it particularly suitable for domains like mental healthcare billing, where anomalies are rare and often unlabeled.

Although pseudo-labeling has shown promise for minority class enrichment, its combination with hybrid learning models is still underexplored. This study addresses this gap by evaluating an LSTM-Transformer hybrid model trained on pseudo-labeled mental healthcare billing data. Beyond fraud detection in mental healthcare, recent studies have highlighted the broader applicability of deep learning and data strategies for anomaly detection. Deshamudre et al.~\cite{deshamudre2023} emphasized the role of structured data preparation and neural architectures in enhancing heart disease prediction accuracy. Similarly, Chatterjee et al.~\cite{chatterjee2023} explored the performance and ethical dimensions of machine learning-based fraud detection in payment systems, highlighting the importance of explainability and domain alignment. These works support the relevance of combining data strategies and model selection when addressing complex, high-stakes anomaly detection problems.

\section{Methodology}
\label{sec:methodology}

\begin{figure}[h!]
  \centering
  \includegraphics[width=0.8\linewidth]{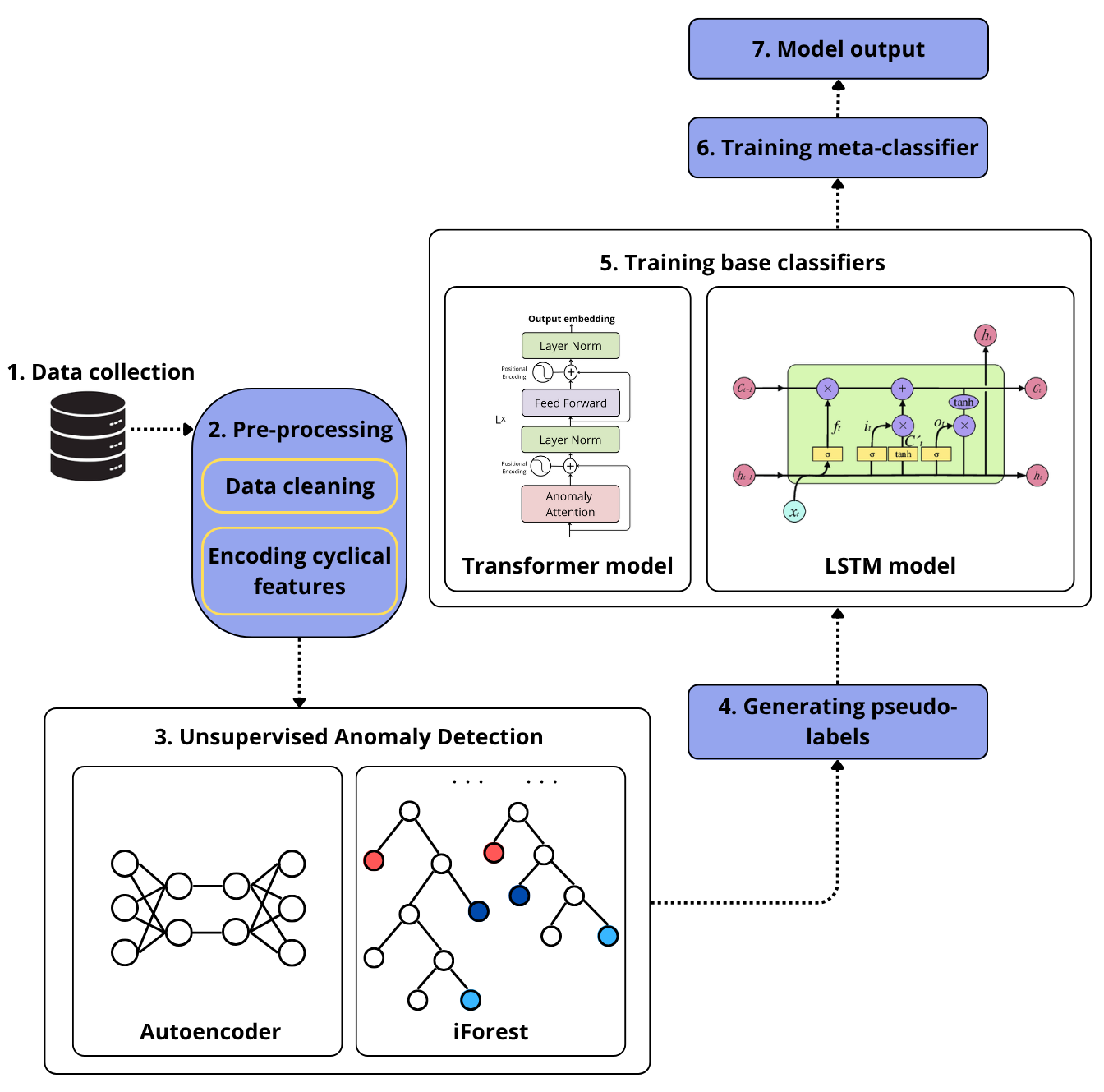}
  \caption{Overview of the hybrid model pipeline for anomaly detection. The process includes data collection, pre-processing, and unsupervised anomaly detection. Detected anomalies generate pseudo-labels for training base classifiers, whose predictions are combined by a meta-classifier to produce the final output.  }
  \label{fig:researchpipeline}
\end{figure}

This study evaluates a hybrid anomaly detection model for imbalanced mental healthcare billing data. It combines an LSTM, a Transformer, and a meta-learner to aggregate their predictions. This section outlines the dataset, pre-processing, pseudo-label generation via unsupervised anomaly detection, model implementation, and performance evaluation methods. Figure \ref{fig:researchpipeline} illustrates the hybrid model pipeline.

\subsection{Datasets}

The datasets used in this study were obtained through a collaboration and comprise two proprietary real-world datasets: one at the declaration-level and the other at the operation-level. Both contain anonymized financial metadata from mental healthcare providers. Each dataset includes binary fraud labels, indicating whether a record has been flagged as fraudulent.

The declaration-level dataset consists of 74,699 billing records across 16 variables from March 2022 to March 2025. The records include invoice amounts, accepted claims, insured coverage, and metadata such as submission dates, payment terms, and early payment requests. One confirmed fraud case was identified before receiving the datasets, involving 349 billing records linked to a single client out of 202 total clients.

The operation-level dataset contains 1,593,647 records across 24 variables, covering operation-level billing records from March 2022 to mid-April 2025. It links individual operations to their corresponding declarations and includes treatment codes, claim rejections, and submission dates. It covers the same 202 clients and 1,799 unique mental healthcare practitioners. The known fraudulent client appears in up to 84 operations tied to a single declaration, with an average of two unique operations per declaration overall.

\subsection{Pre-processing}

With already aligned units, both datasets were cleaned of erroneous entries, normalized, and encoded for cyclical features. NaN values and non-positive billing amounts were removed, resulting in 71,196 entries and 16 columns for the declaration-level dataset and 1,589,951 entries and 24 columns for the operation-level dataset. Other outliers were retained, as they may indicate fraudulent~behavior.

\subsubsection{Encoding cyclical features}

Temporal features were transformed into two dimensions using sine and cosine transformations to capture cyclical patterns. These transformations map each time-related value to a corresponding point in a unit circle, ensuring continuity between the start and end of each cycle. This approach aligns with recent studies on time series data pre-processing for machine learning models \cite{schranz-2020, wilfling-2023}. The date time column was split into year, month, day of week, day, hour, minute, and second. The year variable was retained as a numerical feature since years do not exhibit cyclical behavior. To capture the cyclic nature of these features, they were encoded using the following equations:

\begin{equation}
    \begin{aligned}
        x_{sin} & = \sin(\frac{2 \cdot \pi \cdot x}{\max(T)})\\x_{cos} & = \cos(\frac{2 \cdot \pi \cdot x}{\max(T)})      
    \end{aligned}
\end{equation}

Where \(x\) is the cyclical feature (e.g., hour or day) and \(max(T)\) is the maximum value for a certain cycle (e.g., 24 hours).

\subsubsection{Feature scaling and encoding}

All numerical features were normalized using Min-Max scaling to ensure stable training for gradient-based models, such as LSTMs and Transformers. Furthermore, to enable quantitative representation, all categorical features were one-hot encoded. In both datasets, the payment term (1-6 weeks) and file rejection reason were one-hot encoded, resulting in seven new features.

\subsubsection{Sliding window}

To enable sequential modeling, a fixed-length sliding window approach was applied. Each window contained chronologically ordered features and was labeled by its final timestep. Exploratory data analysis showed cyclical and repetitive billing behavior, supporting the use of sliding windows to capture temporal dependencies. Data was sorted chronologically to preserve temporal order. Sliding windows were extracted from the full dataset without grouping by entity, allowing the model to learn global patterns. Shorter sequences were removed to avoid zero-padding, which could distort learning. A default window size of 100 was used to capture long-range dependencies. For the LSTM and Transformer models trained on pseudo-labeled operation-level data, a window size of 50 was applied. This reduced dimensionality and improved performance based on empirical testing. In contrast, the original-label model used size 100, as extreme class imbalance required the model to leverage more temporal context to detect subtle~anomalies.

\subsection{Unsupervised anomaly detection}

The original binary fraud labels were excluded from all modeling steps due to extreme class imbalance, with only 0.4\% of instances labeled as anomalous. To address this, this study adopted an unsupervised pseudo-labeling approach based on anomaly techniques. This strategy identifies outliers from structural deviations present in the data without relying on labels. All pseudo-labels were derived from the original datasets, without introducing any synthetic anomalies. Two distinct unsupervised models were applied to generate pseudo-labels: Isolation Forest (iForest) and Autoencoder (AE). Both models were trained on the declaration-level and operation-level datasets, enabling comparison of their effectiveness. The pseudo-codes for the iForest and AE pseudo-labeling models are presented in Algorithm \ref{alg:iforest} and Algorithm \ref{alg:autoencoder}. Label quality was visually validated using t-SNE and density plots, and agreement between the two models was inspected to assess consistency.

\subsubsection{Isolation Forest}
\sloppy
IForest, known for its effectiveness in anomaly detection and scalability on large datasets, was used for pseudo-labeling. IForest is particularly well-suited for high-dimensional data, where it isolates anomalies by partitioning data points within an ensemble of random isolation trees. Anomalies are expected to be isolated more quickly, which results in shorter average path lengths. IForest is robust to outliers and can effectively isolate anomalies even in complex and non-linear datasets \cite{liang2022self, xu-2023, Bansal-2024}. Instances received anomaly scores between -1 and 1, where negative values indicate outliers and positive values indicate inliers. Data points originally labeled as non-fraud but exceeding a defined anomaly score threshold \(\theta\) were assigned a pseudo-label of 1. 

\begin{algorithm}[htbp]
\caption{Isolation Forest for pseudo-labeling}
\label{alg:iforest}
\KwIn{$X = \{x_1, x_2, \dots, x_n\}$ - Input dataset with $n$ instances}
\KwIn{$T$ - Number of isolation trees}
\KwIn{$c$ - Contamination rate}
\KwOut{Pseudo-labeled data with labels $y_i$}

\textbf{Initialize} empty forest $F \leftarrow \emptyset$ \\
\For{$t = 1$ \KwTo $T$}{
    Sample random subset $X_t$ from $X$ \\
    Build isolation tree $T_t$ on $X_t$ \\
    Add $T_t$ to $F$ \\
    \Return $F$
}
\For{each $x_i \in X$}{
    Compute average path length $h(x_i)$ across all trees \\
    Calculate anomaly score $S(x_i)$ based on $h(x_i)$
}
\textbf{Sort} all instances $x_i$ by $S(x_i)$ in descending order\\

\textbf{Calculate} threshold $\theta$ as the anomaly score of the instance at position $\lfloor c \cdot n \rfloor$\\

\For{each $x_i \in X$}{
    \eIf{$S(x_i) < \theta$}{
        Assign pseudo-label $y_i = 1$
    }{
        Assign pseudo-label $y_i = 0$
    }
}
\end{algorithm}

\subsubsection{Autoencoder}

A limitation of iForest is its reliance on a tree-based approach, which can struggle to capture more complex anomaly patterns, as each isolation operation considers only one feature at a time \cite{xu-2023}. To address this limitation, AE has been applied as a separate method for anomaly detection and pseudo-labeling. AE is an unsupervised artificial neural network consisting of an input layer, a hidden layer, and an output layer. It operates through an encoder-decoder structure, where the encoder reduces the initial input data by mapping it to a hidden representation with a smaller dimension. The decoder reconstructs the original input from this compressed form. The main objective of an AE is to minimize the reconstruction error \cite{abhaya-2023}. Anomalous instances, which differ from the majority of the data, typically exhibit higher reconstruction errors due to their dissimilarity to the learned normal patterns. By generating pseudo-labels based on the reconstruction error, AE offers a complementary method to iForest for capturing more complex anomaly patterns that iForest might miss.

\begin{algorithm}[htbp]
\caption{Autoencoder for pseudo-labeling}
\label{alg:autoencoder}
\KwIn{$X = \{x_1, x_2, \dots, x_n\}$ - Input dataset with $n$ instances}
\KwIn{$\text{AE model} = \text{Encoder, Decoder}$}
\KwIn{$\theta_{error}$ - Threshold for reconstruction error}
\KwOut{Pseudo-labeled data with labels $y_i$}

\For{each $x_i \in X$}{
    \textbf{Encode} $h_i = f_{\text{encoder}}(x_i)$ - Map input to hidden representation\\
    \textbf{Decode} $\hat{x}_i = f_{\text{decoder}}(h_i)$ - Reconstruct the input from hidden representation\\
    \textbf{Calculate reconstruction error} $e_i = \|x_i - \hat{x}_i\|^2$ 
}

\For{each $x_i \in X$}{
    \eIf{$e_i > \theta_{error}$}{
        Assign pseudo-label: $y_i = 1$ 
    }
    {
        Assign pseudo-label: $y_i = 0$ 
    }
}
\end{algorithm}

\subsection{Model architectures}

This subsection describes the LSTM, Transformer, and hybrid model architectures for anomaly detection, leveraging the pseudo-labels generated by the unsupervised anomaly detection models. The LSTM and Transformer models were selected for their ability to model sequential and long-range dependencies, with a hybrid model combining their strengths via stacking. The models were trained using sliding window sequences as input, with pseudo-labels from iForest and AE as target outputs.

\subsubsection{LSTM model}

LSTMs are well-suited for capturing long-term dependencies within sequential data, making them effective for anomaly detection in billing records. This stems from their use of memory cells and gating mechanisms, which mitigate the vanishing gradient problem found in traditional RNNs \cite{ileberi-2024,kumar-2019,barrera-animas-2021}. The model used in this study consists of two LSTM layers with 64 hidden units each, followed by a fully connected output layer. A single-layer variant was used for the original-label model on the declaration-level dataset, as it demonstrated better performance during training.

\subsubsection{Transformer model}

\begin{figure}[h!]
  \centering
  \includegraphics[width=0.3\linewidth]{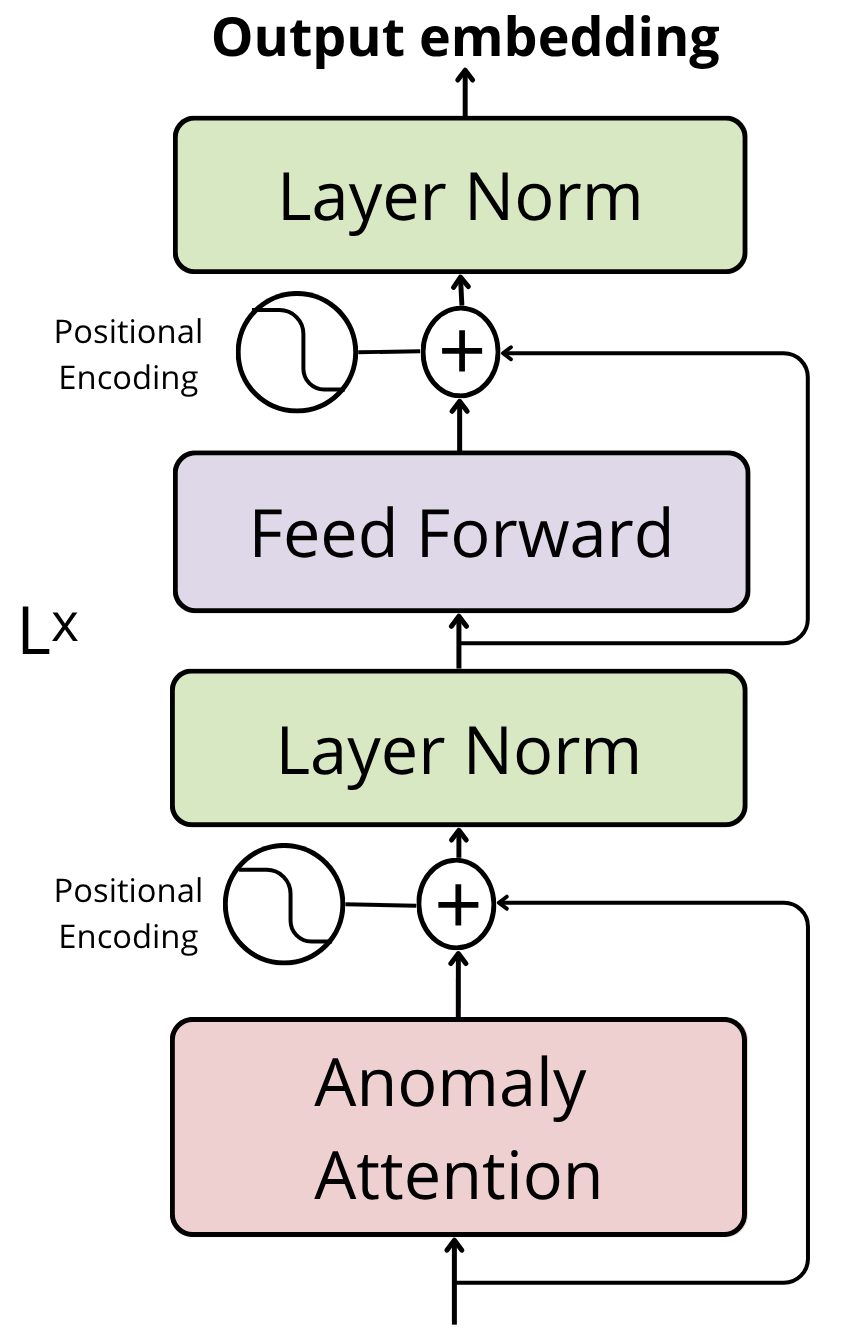}
  \caption{Adapted Anomaly Transformer architecture. The model employs Anomaly-Attention layers and feed-forward networks, each with residual connections and layer normalization. The reconstruction component is replaced with a classification head after the final encoder block.}
  \label{fig:transformeranomalyalg}
\end{figure}

Transformers were chosen for their ability to capture long-range dependencies without relying on sequential processing. Their self-attention mechanism enables the model to determine the importance of each input element, which is useful for identifying subtle, temporal anomalies in billing records. This study adapted the encoder architecture from the Anomaly Transformer by Xu et al. \cite{xu2022anomaly}, originally developed for unsupervised time series anomaly detection. The adapted model includes three encoder layers, each containing multi-head self-attention with eight attention heads, positional encoding, and a feedforward network with a hidden size of 512. Instead of reconstruction, a classification head consisting of a linear layer applied to the final timestep embedding was added. The Anomaly Transformer extends self-attention by modeling temporal dependencies via learned prior and series associations. The prior association uses a learnable Gaussian kernel to emphasize local dependencies, allowing the model to detect anomalies in nearby time windows. The series association captures broader patterns by learning attention weights across the full sequence, enabling the detection of long-span anomalies. The adapted Anomaly Transformer architecture is shown in Figure \ref{fig:transformeranomalyalg}.

\subsubsection{Hybrid model}

To combine the strengths of the LSTM and Transformer models, a stacking ensemble approach was used. Unlike bagging or boosting, which typically use homogeneous base learners and deterministic aggregation, stacking allows the integration of heterogeneous architectures. As a non-deterministic method, stacking combines predicted class labels from independently trained base models and uses them as input features for a meta-learner to produce the final classification. By doing so, stacking can potentially minimize the prediction error, reduce noise, and explore the combined power of different models, as demonstrated by Cao et al. \cite{cao-2024}. Logistic Regression (LR) was selected as the meta-learner due to its simplicity, robustness, and efficiency in low-dimensional spaces, as it only processes the base models' outputs. A recent study by Shafieian and Zulkernine \cite{shafieian-2022} further supports this approach, showing that stacking outperforms bagging and boosting in detecting anomalies within complex network environments.

\subsection{Experimental setup}

A series of experiments evaluated the LSTM, Transformer, and hybrid models. All models were trained on both the original-label and their corresponding pseudo-labeled datasets, resulting in six variants. Unsupervised anomaly detection was performed locally, while model training and evaluation were run on the Snellius supercomputer \footnote{https://www.surf.nl/diensten/rekenen/snellius-de-nationale-supercomputer}. The hyperparameter settings for all models are provided in Appendix \ref{sec:hyperparameters}.

\subsubsection{Unsupervised anomaly detection}

An 80/20 train-test split was applied before training the unsupervised models. With scarce ground-truth labels and a single fraud case unlikely to represent the full range of anomaly patterns, traditional evaluation metrics were not applicable. Instead, anomaly score distributions and label assignment behavior were analyzed. Both iForest and AE models were optimized using random search and executed multiple times to account for variability. The best performing configurations were selected based on mean absolute deviation for iForest and negative mean absolute error for the AE.

For iForest, the contamination threshold was set at 1.6\% based on domain expertise. Instances exceeding this threshold received a pseudo-label of 1. The AE consisted of an encoder and decoder, each with two fully connected layers using ReLU activation. The encoder included two hidden layers, and the decoder mirrored this structure, ending with an output layer using Sigmoid activation. Early stopping was applied during training with a patience of eight epochs. Anomalies were identified using reconstruction error based on mean absolute error, applying the same 1.6\% threshold used for iForest and guided by domain knowledge.

\subsubsection{Baseline and hybrid models}

LSTM and Transformer models served as baselines. All models used a 60/20/20 train-validation-test split, applied in chronological order to prevent temporal leakage. Validation sets were used for model tuning, hyperparameter optimization, and early stopping. Test sets were held out for final evaluation. 

LSTM and Transformer models were implemented in PyTorch and trained separately on each of the six dataset variants. Both models used a weighted binary cross-entropy loss, with dataset-specific positive weights, and were optimized using Adam. Training ran up to 50 epochs, with early stopping based on validation loss. Classification thresholds were selected using the F1-score from the precision-recall curve or set to 0.5 if no optimum was found. To ensure reliable results, all models were run multiple times using the optimal threshold.

LSTM models used a dropout rate of 0.5 to reduce overfitting, except for the original-label declaration-level dataset, where dropout was not applicable. Early stopping was applied after seven epochs of stagnant validation loss, and a ReduceLROnPlateau scheduler reduced the learning rate by 0.5 after 5 epochs without improvement. Transformer models were trained without dropout but used a custom learning rate scheduler that halved the learning rate after each epoch to stabilize convergence. Early stopping was triggered after 10 epochs without improvement. 

The hybrid model used LR to combine the label predictions from LSTM and Transformer models. For each dataset variant, predicted class labels from both models on validation and test sets were used as feature sets. LR was trained on validation features, with the two models' validation labels as the target. Class weights were set to 'balanced' to adjust for class imbalance.

\subsection{Model interpretation and explainability}

SHapley Additive exPlanations (SHAP) \cite{lundberg-2017} was applied to improve the interpretability of complex deep learning models. Based on principles from game theory, SHAP assigns Shapley values to individual features, capturing their contribution to each prediction. This provides a clear understanding of the importance of input features in the decision-making process of deep learning models.

\subsection{Performance evaluation metrics}

To evaluate the performance of the hybrid and baseline models, precision, recall, and F1-score were used as the primary performance metrics. These metrics were calculated for each model across both original and pseudo-labeled datasets. Accuracy was included for completeness but interpreted with caution, as it can overestimate performance by favoring the majority class. For example, a naive classifier could achieve 99\% accuracy simply by labeling all samples as non-anomalous. Precision measured the proportion of correctly classified anomalous instances, while recall measured the proportion of actual anomalous instances correctly identified by the classifier. The F1-score provided the harmonic mean of recall and precision. Additionally, McNemar's test \cite{mcnemar-1947} was applied to assess whether the difference in classification errors between two models trained on the same pseudo-labels was statistically significant.

\section{Results}
\label{sec:results}


This section evaluates the anomaly detection performance of the baseline and hybrid models on both the declaration- and operation-level datasets. The results demonstrate the effectiveness of the proposed models in handling highly imbalanced and skewed data.

\subsection{Unsupervised anomaly detection}

To address SQ1, the quality of the pseudo-labels generated by the iForest and AE models is assessed using t-SNE embeddings and density plots of anomaly scores and reconstruction errors. In both datasets, AE-based pseudo-labels show clearer separation between normal and anomalous instances, indicating stronger anomaly isolation. IForest performs comparatively well on the operation-level dataset, where anomalies are grouped more closely together.

\subsubsection{T-SNE embeddings}

\begin{figure}[h!]
  \centering
  \includegraphics[width=0.7\linewidth]{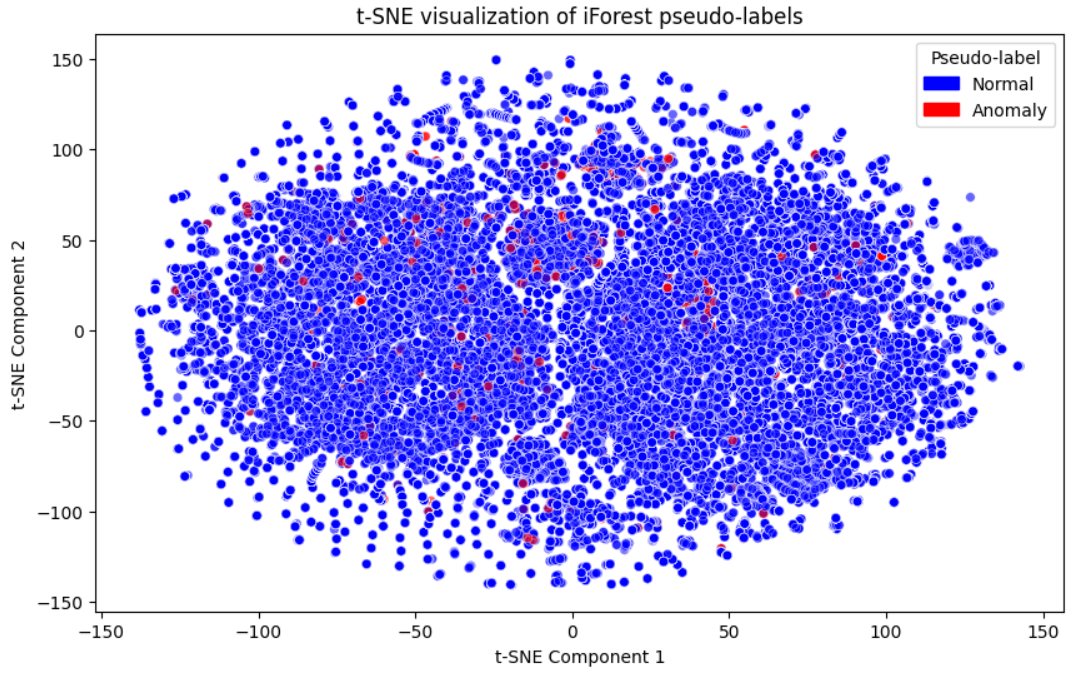}
  \caption{T-SNE plot of Isolation Forest pseudo-labels for the declaration-level dataset. Anomalous instances are scattered with limited separation from normal instances.}
  \label{fig:iforesttsne}
\end{figure}

\begin{figure}[h!]
  \centering
  \includegraphics[width=0.7\linewidth]{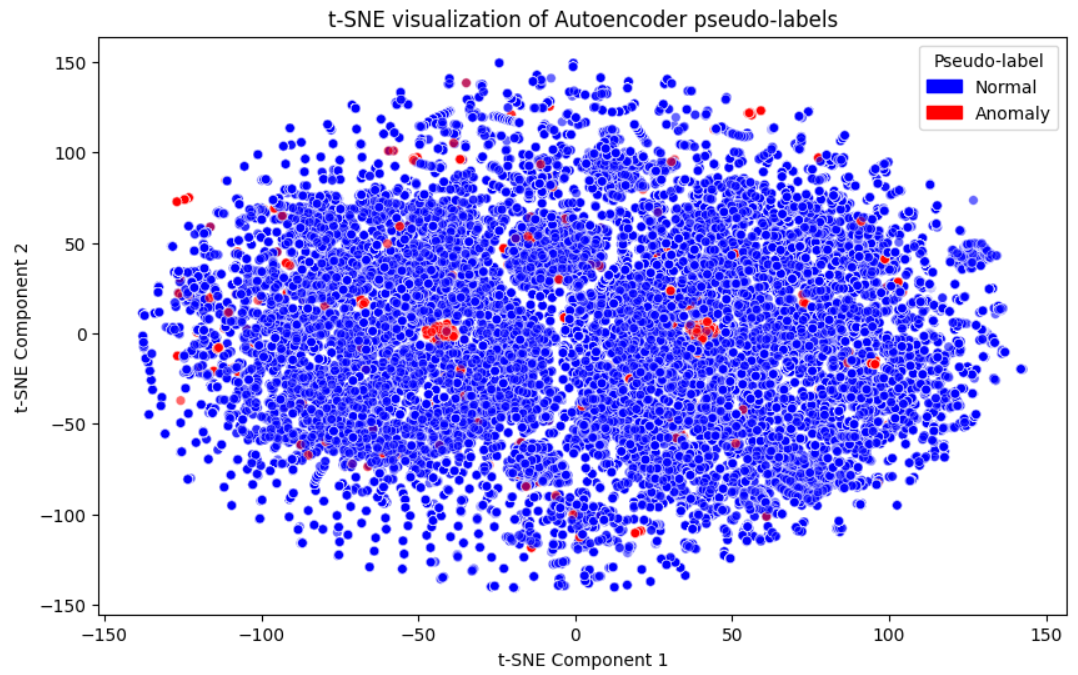}
  \caption{T-SNE plot of Autoencoder pseudo-labels for the declaration-level dataset. Anomalous instances form multiple localized clusters, indicating clearer separation.}
  \label{fig:autoencodertsne}
\end{figure}

The operation-level dataset was subsampled to 50,000 instances due to computational limits. For the declaration-level dataset, the iForest t-SNE plot (Figure \ref{fig:iforesttsne}) shows scattered anomalies with limited separability. In contrast, the AE t-SNE plot (Figure \ref{fig:autoencodertsne}) reveals clearer separation, with anomalies forming distinct localized clusters. On the operation-level dataset (Appendix \ref{sec:appUnsupervisedanomalydetection}) both models show improved clustering. AE anomaly clusters appear near the lower edge of the embedding space, while iForest anomalies form clusters in the lower central region. This indicates that AE better captures the underlying structure of anomalies, making it possible to identify deviant patterns more effectively.

\subsubsection{Anomaly distribution}
\begin{figure}[h!]
  \centering
  \includegraphics[width=0.7\linewidth]{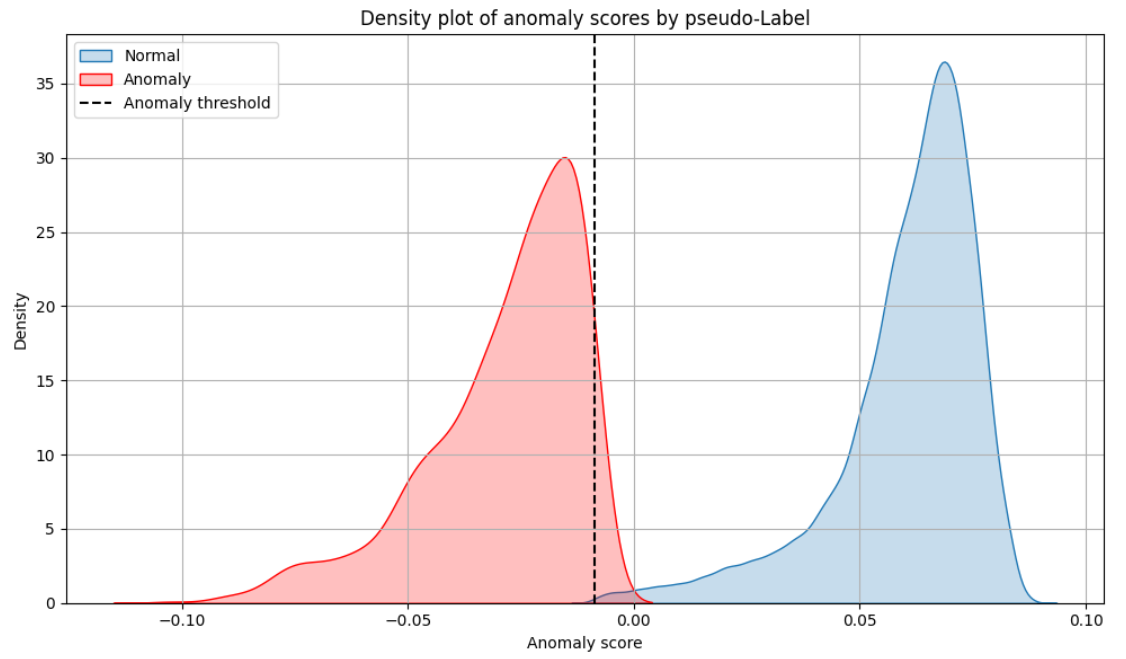}
  \caption{Density plot of Isolation Forest anomaly scores for the declaration-level dataset. Normal instances peak around 0.05, while anomalies peak near -0.025, showing clear separation with a decision threshold at -0.01.}
  \label{fig:iforestdensity}
\end{figure}

\begin{figure}[h!]
  \centering
  \includegraphics[width=0.7\linewidth]{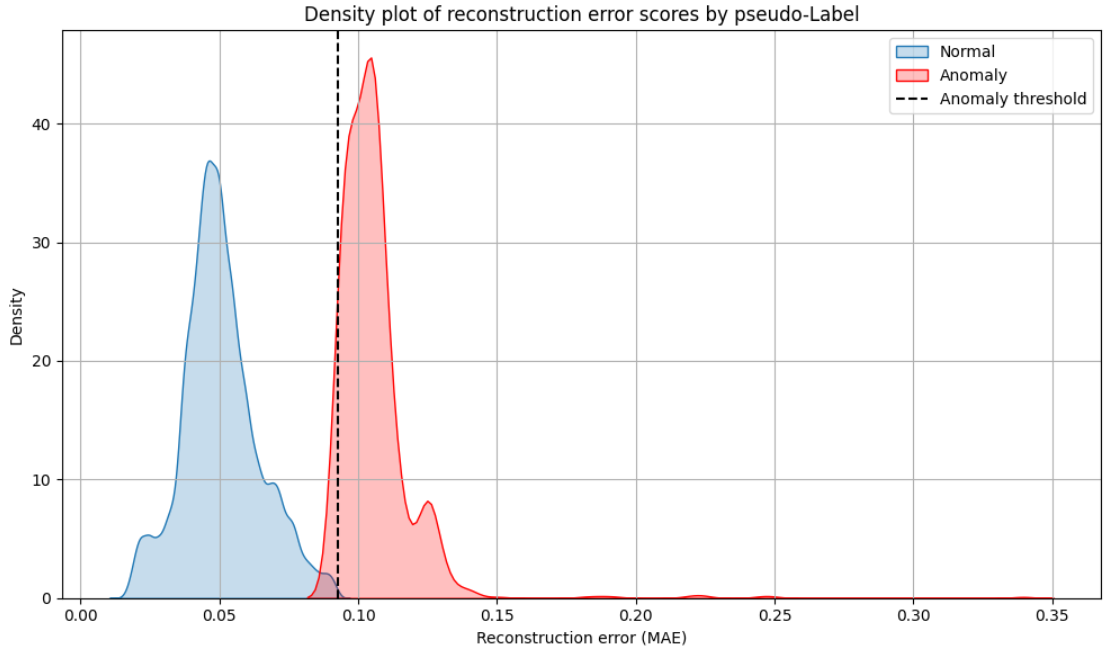}
  \caption{Density plot of Autoencoder reconstruction errors for the declaration-level dataset. Normal instances peak at 0.05, while anomalies peak around 0.10. The threshold at 0.08 indicates strong separability between the two groups.}
  \label{fig:autoencoderdensity}
\end{figure}

The density plots reveal the distribution of anomaly scores (iForest) and reconstruction errors (AE) by pseudo-labels. For the declaration-level dataset, the iForest distribution (Figure \ref{fig:iforestdensity}) shows a clear separation, with normal instances peaking at 0.05 and anomalies at -0.025, with a threshold of -0.01. The operation-level plot (Appendix \ref{sec:appUnsupervisedanomalydetection}) follows a similar pattern, with even clearer separation. For the AE model, the declaration-level errors (Figure \ref{fig:autoencoderdensity}) show normal instances peaking at 0.05 and anomalies peaking at 0.10, with a threshold of 0.08. The operation-level distribution (Appendix \ref{sec:appUnsupervisedanomalydetection}) is more spread out. However, normal instances form two peaks around 0.10 and 0.11, with limited overlap, indicating that most normal instances fall within this range. Across both datasets, iForest anomaly scores are left-skewed, reflecting that anomalies tend to receive lower scores. In contrast, AE reconstruction errors are right-skewed, indicating that anomalies typically have larger reconstruction errors compared to normal instances. These patterns align with expectations for imbalanced data, where rare instances appear in the distribution tails.

To assess the alignment between iForest and AE, their pseudo-label agreement was examined. In both datasets, pseudo-label agreement exceeds 97\% for normal instances (label 0). In contrast, only a small fraction of both datasets are consistently labeled as anomalies (label 1). Approximately 2.7\% of instances show disagreement, where one model labels an instance as anomalous while the other does not. This limited overlap in positive labels highlights fundamental differences in how the models define anomalies, consistent with the patterns observed in the t-SNE embeddings. Full label agreement statistics are provided in Appendix \ref{sec:appUnsupervisedanomalydetection}.

To better understand the identified anomalies, the top ten highest-scoring instances were selected for each model, allowing manageable analysis while capturing key model differences. In the declaration-level dataset, one client appears three times in the iForest top ten and five times in the AE top ten, with three declared files flagged by both. In the operation-level dataset, one client appears six times in the iForest top ten, with each instance corresponding to the same declaration file. AE identifies this same client twice, but for different files, and highlights another client six times across three different declaration files. These patterns suggest that iForest focuses on repeated structural anomalies within a single context, while AE tends to identify a wider spread of anomalies across different files or entities. This implies that iForest may be better suited for detecting recurring, structural anomalies, while AE may be more effective at uncovering diverse and context-specific anomalies. Feature statistics are included in Appendix \ref{sec:appUnsupervisedanomalydetection}.

\subsection{Model performance}

In relation to SQ2, this subsection compares the anomaly detection performance of the baseline models and hybrid models trained on the pseudo-labeled and original-label data. Model performance is evaluated using accuracy, precision, recall, and F1-score. Statistical differences in classification errors are assessed using McNemar's test. The results from the test set, reflecting model generalization, are discussed. Table \ref{tab:evaldbasedtest} and Table \ref{tab:evalobasedtest} present the results for the declaration-level and operation-level datasets. Validation set results are provided in Appendix \ref{sec:evalmetricsappendix} for completeness.

\subsubsection{McNemar's test}
McNemar's test reveals that the differences in classification errors between the LSTM and Transformer models are statistically significant under the same pseudo-labeling conditions. This holds for both the declaration-level and operation-level datasets, using iForest and AE pseudo-labels, across validation and test sets (\textit{p }< 0.05). This suggests that model architecture has a substantial impact on anomaly detection behavior, even when trained on identical pseudo-labels. Detailed contingency tables are included in Appendix \ref{sec:evalmetricsappendix}.

\subsubsection{IForest-based models}

Among the baseline models, iForest-based models perform consistently well on both datasets. On the declaration-level dataset, the iForest LSTM achieves the highest scores across all metrics. The iForest Transformer performs reasonably but shows lower recall, suggesting a tendency to miss true anomalies. The hybrid iForest-based model demonstrates moderate improvement over the Transformer, though the iForest LSTM remains the strongest performer overall. On the operation-level dataset, the iForest LSTM again outperforms all models, achieving the best results on both validation and test sets. In contrast, the iForest Transformer shows a significant drop in test performance relative to its validation results, reflecting poor generalization. Except for the iForest LSTM, all models show substantial drops in generalization, indicating overfitting to the training data. The hybrid variant achieves the highest recall but still lags behind the iForest LSTM in overall accuracy, precision, and F1-score.
 
\subsubsection{AE-based models}

AE-based models show strong performance on the declaration-level dataset but struggle to generalize to more complex data. Both the AE LSTM and AE Transformer perform reasonably well, though with lower recall. Notably, the AE LSTM achieves higher precision and recall on the test set than on the validation set. This can be the result of a higher anomaly proportion in the test set compared to the validation set, creating more opportunities for the model to correctly identify these anomalies. The hybrid AE-based model outperforms its baselines in recall and F1-score but with a slight reduction in precision. In contrast, on the operation-level dataset, AE-based models perform poorly. Both the AE Transformer and AE LSTM show severe drops in performance, with results below random guessing. While the AE LSTM performs moderately on the validation set, its test performance drops significantly, indicating overfitting. The hybrid variant also fails to improve performance, further demonstrating the difficulty AE-based models have in generalizing to more complex data.

\subsubsection{Generalization and dataset complexity}

A key insight in addressing SQ3 is the variation in model performance across datasets. While iForest-based models, particularly the iForest LSTM and its hybrid variant, consistently outperform others, most models show notable performance drops on the operation-level setting. This suggests that the operation-level dataset may be too complex for the models to reliably distinguish anomalies and produce consistent results. Moreover, hybrid models tend to increase recall at the cost of precision. These patterns underscore the importance of selecting models based on the specific anomaly detection task and highlight the potential of hybrid models to balance sensitivity and generalization.

\subsubsection{Original-label based models}

Across both datasets, original-label models consistently fail to detect anomalies. All baseline and hybrid variants score zero across evaluation metrics. While some models show minor results on the validation sets, their overall performance remains poor. In several test splits, no labeled anomalies were present, preventing the models from distinguishing normal from anomalous instances. This consistent failure likely reflects extreme class imbalance, which leaves the models unable to learn meaningful patterns from the minority class. It suggests that model architecture alone is not sufficient without label enrichment, even when adopting deep learning models. These results confirm the limitations of using standard supervised learning techniques while dealing with highly imbalanced data. As noted by Benchaji et al. \cite{benchaji-2021}, traditional classifiers often fail to detect anomalies when normal instances dominate.

\begin{table}[ht]
\centering
\caption{Model performance on the declaration-level dataset based on test set metrics. IForest LSTM achieves the highest performance across all metrics. Models trained on original labels fail to detect anomalies.}
\small 
\begin{tabular}{lcccc}
\toprule
\textbf{Classifiers} & \textbf{Accuracy} & \textbf{Precision} & \textbf{Recall} & \textbf{F1}\\
\midrule
\multicolumn{5}{l}{\textbf{Baseline models}} \\
AE Transformer & 0.987& 0.940& 0.556& 0.698\\
iForest Transformer& 0.994& 0.806& 0.876& 0.840\\
Original-label Transformer& 0.000 & 0.000& 0.000 & 0.000 \\
AE LSTM & 0.991 & 0.912& 0.705& 0.795\\
iForest LSTM& \textbf{0.999}& \textbf{0.995}& \textbf{0.963}& \textbf{0.959}\\
Original-label LSTM& 0.520& 0.000& 0.000 & 0.000 \\
\midrule
\multicolumn{5}{l}{\textbf{Hybrid models}} \\
Hybrid AE based& 0.991& 0.901& 0.715&0.798\\
Hybrid iForest based& 0.996& 0.818& \textbf{0.963}&0.884\\
Hybrid original-label based& 0.000 & 0.000& 0.000 & 0.000 \\
\bottomrule
\end{tabular}
\label{tab:evaldbasedtest}
\end{table}

\begin{table}[ht]
\centering
\caption{Model performance on the operation-level dataset based on test set metrics. IForest LSTM achieves the best overall balance of precision and recall. The hybrid iForest-based model improves recall but with reduced precision. All original-label models fail to identify anomalies.}
\small 
\begin{tabular}{lcccc}
\toprule
\textbf{Classifiers} & \textbf{Accuracy} & \textbf{Precision} & \textbf{Recall} & \textbf{F1}\\
\midrule
\multicolumn{5}{l}{\textbf{Baseline models}} \\
AE Transformer & \textbf{0.997}& 0.216& 0.099& 0.136\\
iForest Transformer& 0.977& 0.183& 0.298& 0.226\\
Original-label Transformer& 0.735 & 0.000 & 0.000 & 0.000 \\
AE LSTM & 0.995& 0.150& 0.310& 0.202\\
iForest LSTM& 0.996& \textbf{0.918}& 0.699& \textbf{0.794}\\
Original-label LSTM& 1.0& 0.000 & 0.000 & 0.000 \\
\midrule
\multicolumn{5}{l}{\textbf{Hybrid models}} \\
Hybrid AE based& 0.995 & 0.148& 0.310&0.200\\
Hybrid iForest based& 0.976& 0.279& \textbf{0.744}&0.406\\
Hybrid original-label based& 1.0& 0.000& 0.000 & 0.000 \\
\bottomrule
\end{tabular}
\label{tab:evalobasedtest}
\end{table}

\subsection{SHAP}

\begin{figure}[h!]
  \centering
  \includegraphics[width=0.7\linewidth]{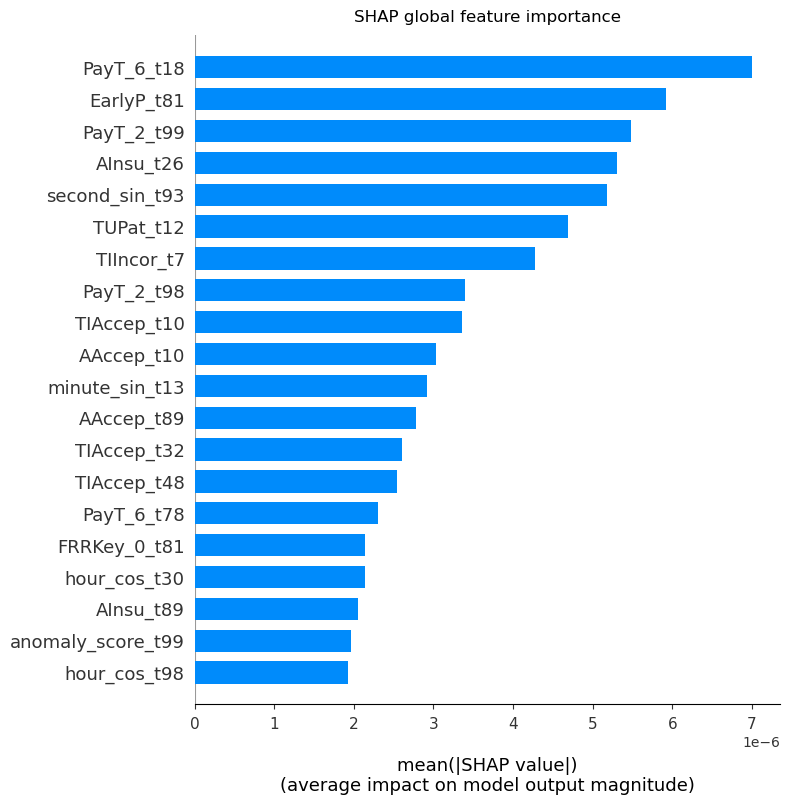}
  \caption{SHAP summary bar plot, demonstrating the top 20 most influential features in the iForest LSTM model. Features related to payment timing and opting for early payment dominate feature importance.}
  \label{fig:shapiforestlstmbar}
\end{figure}

\begin{figure}[h!]
  \centering
  \includegraphics[width=0.6\linewidth]{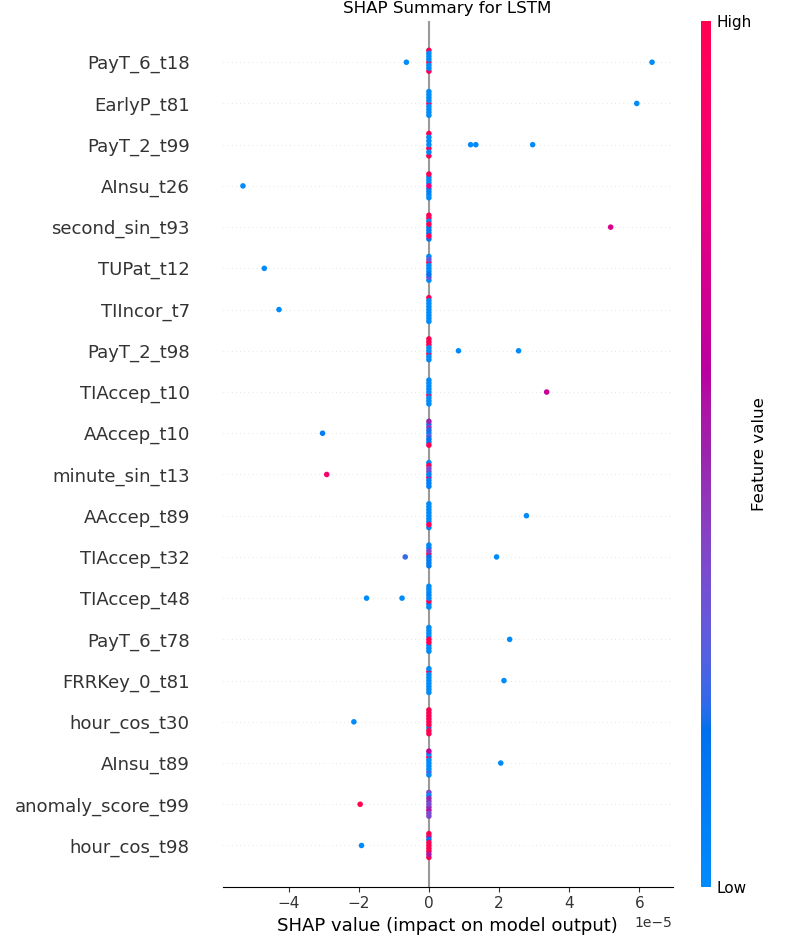}
  \caption{SHAP summary plot, demonstrating the distribution and direction of feature impact in the iForest LSTM model. High and low feature values influence anomaly predictions in a context-dependent manner.}
  \label{fig:shapiforestlstm}
\end{figure}

Figure \ref{fig:shapiforestlstmbar} and Figure \ref{fig:shapiforestlstm} present a SHAP-based interpretability analysis of the iForest LSTM model. As the best-performing model, it was selected for explainability evaluation. The SHAP bar plot shows that the payment term \texttt{(PayT\_6\_t18)} is the most important feature, indicating that earlier time step values strongly influence anomaly predictions. The SHAP summary plot reveals that both high and low feature values impact predictions, suggesting a context-dependent role. Other top features, such as \texttt{EarlyP\_t81} (early payments) and \texttt{PayT\_2\_t99}, show that lower feature values are associated with higher SHAP values, increasing the likelihood of the model predicting an anomaly. Features like \texttt{hour\_cos\_t30} show low importance, indicating limited relevance of cyclical time. Additional SHAP plots for baseline models are included in Appendix \ref{sec:shapappendix}.

\section{Discussion}
\label{sec:discussion}

This section reflects on the methods employed, evaluating their strengths and limitations, and outlines directions for future research. Moreover, ethical considerations are discussed, particularly the implications of applying deep learning in sensitive domains like mental healthcare billing. Lastly, it outlines opportunities for future research.

\subsection{Analysis of findings}

The iForest LSTM model achieved the highest performance across both datasets. This aligns with prior work by Benchaji et al. \cite{benchaji-2021}, who demonstrated that LSTMs are effective in capturing long-term dependencies in sequential financial data. The success of the iForest LSTM model in this study further supports the applicability of sequential deep learning models in the context of billing fraud. Transformer models, while promising in other anomaly detection domains \cite{tang-2024, yu-2024, xu2022anomaly}, showed less stable performance in this study. This may be due to the complexity of tuning attention-based architectures and their sensitivity to noise and variation in the input features. McNemar's test confirmed significant classification error differences between LSTM and Transformer models, even when trained on identical pseudo-labels. This highlights that the architecture choice, combined with the labeling strategy, has a substantial impact on prediction outcomes. Only the iForest-based hybrid model yielded more robust outputs, though not surpassing the iForest LSTM baseline. This contrasts with findings by Ileberi and Sun \cite{ileberi-2024}, who reported that hybrid architectures outperformed individual learners in credit card fraud detection. A possible explanation is their use of the European Credit Card Dataset from 2013. Given its age, the dataset may not reflect current fraud patterns or the complexity of domain-specific billing data, limiting the generalizability of their results. The complete failure of the original-label models to detect anomalies further confirms the findings of Kulatilleke \cite{kulatilleke-2022}, who emphasized the limitations of supervised learning under extreme class imbalance. In contrast, the use of pseudo-labeling improved model performance, aligning with the findings of Al-Lahham et al. \cite{al-lahham-2023} and Im et al. \cite{im-2024}, who showed that unsupervised labeling can be effective when labeled anomalies are limited or unavailable.

Although t-SNE visualizations showed greater separability for pseudo-labels generated by the AE, models trained on iForest pseudo-labels consistently performed better. AE pseudo-labels may retain noise and fail to effectively distinguish anomalies from normal instances. This highlights that visual interpretability does not always translate to practical utility. Future work could explore ensemble methods to combine the strengths of both approaches.

\subsection{Limitations}

While this study demonstrates promising results, several limitations must be acknowledged. First, the use of proprietary mental healthcare billing data limits reproducibility. Due to privacy and contractual restrictions, the datasets are not publicly available, which constrains replication of the findings. Moreover, since the datasets originate from a single healthcare declaration source and focus exclusively on mental healthcare, the model is influenced by specific billing practices, domain-specific policies, and fraud patterns. Applying it in other domains or contexts may therefore require retraining to accommodate different data characteristics.

A further consideration involves how the model's output should be interpreted. Since the labeling process relies on unsupervised learning rather than confirmed fraud cases, some of the detected anomalies may represent unusual but legitimate billing behavior. This underscores the need for expert review to assess flagged anomalies. For instance, Ding et al. \cite{ding-2023} proposed the ALARM framework, which integrates a human-in-the-loop approach into the anomaly detection pipeline by incorporating iterative expert review and explanation mechanisms. Adopting similar strategies in mental healthcare billing could help reduce false positives and ensure that automated detection remains aligned with domain expertise. Additionally, while SHAP offers partial insight into the models' decision-making, it provides only a limited understanding of their underlying reasoning. As such, deploying these models in high-stakes environments should be approached with caution and transparency.
\subsection{Ethical and practical considerations}

The application of deep learning models for anomaly detection in mental healthcare billing raises ethical and practical considerations. While these models can flag potential fraud cases that might otherwise go unnoticed, the trade-off between precision and recall is crucial in this domain. Favoring high recall would mean detecting more possible fraud cases and may be prioritized over precision to minimize financial losses. However, this could lead to increased false positives, misclassifying legitimate billing and risking providers' reputations. Since anomalies do not always indicate fraud, domain expertise is vital to interpret outputs and distinguish errors from real fraud cases. Moreover, the black-box nature of deep learning models requires a human-in-the-loop approach to ensure transparency and accountability. Model outputs should prompt expert review, not automated decisions, and patient privacy should always be prioritized over the pursuit of performance metrics.

Another constraint relates to the computational demands of the models. Although hybrid models showed some performance gains, training deep learning models required significant resources and time. This could hinder adoption in settings without access to high-performance computing infrastructures. Transformers, in particular, pose scalability challenges due to their quadratic memory complexity. Practical deployment may require model compression, optimization, or less computationally expensive architectures to ensure feasibility.

\subsection{Future work}

Future research could assess whether the hybrid model generalizes to other domains, such as dentistry or insurance fraud. This would clarify whether the performance gains are generalizable or specific to mental healthcare data. Another key direction is improving model performance on complex data like the operation-level dataset. In this study, performance dropped significantly, especially for the Transformer models. Potential directions include exploring methods to handle this complexity or more targeted feature engineering informed by domain knowledge. While this study implemented a two-layer LSTM, other recurrent configurations were not explored. Future research could evaluate different architectures, such as bidirectional LSTMs, as well as variations in hidden units and sequence lengths, to assess their effect on anomaly performance. Furthermore, future work should consider the trade-offs between model complexity and performance, as this study demonstrated that increased complexity does not always yield better results. Finally, while unsupervised pseudo-labeling improved results, it could be interesting to experiment with different synthetic anomaly generation techniques to further enhance pseudo-labeling quality. 

\section{Conclusion}
\label{sec:conclusion}

This study addressed the challenge of anomaly detection in mental healthcare billing, a domain marked by class imbalance and fraud risks. It investigated whether a hybrid LSTM-Transformer model could enhance anomaly detection in imbalanced billing data and evaluated the role of pseudo-labeling in improving model performance. The iForest LSTM model outperformed both baseline and hybrid models, indicating that simpler sequential models are more effective in this setting. The hybrid iForest-based model, though not surpassing iForest LSTM, demonstrated promising results by improving the standalone Transformer, AE, and supervised models. Pseudo-labeling, particularly with iForest, significantly enhanced model performance over training on original labels, confirming its value in settings with extreme label scarcity.

This research contributes to the field of unsupervised anomaly detection by validating the effectiveness of pseudo-labeling in the mental healthcare billing domain. It highlights the trade-offs between model complexity and performance, suggesting that simpler architectures may be more suitable for practical deployment. Future work could focus on model variants and synthetic anomaly generation to address data imbalance and improve pseudo-label quality, while also evaluating generalizability across domains.

\bibliographystyle{ACM-Reference-Format}


\newpage

\onecolumn

\appendix
\begin{appendices}

\section{ Hyperparameters}
\label{sec:hyperparameters}

For the sake of reproducibility, the tested and final hyperparameter settings for all models, including the unsupervised anomaly detection, are reported below. The tables are separated by model type and dataset-level. Given the computational costs of the LSTM and Transformer models, exhaustive methods such as grid search or random search were not feasible. Instead, hyperparameters, such as the learning rate, batch size, and patience, were adjusted based on performance trends during validation.

\begin{table}[ht]
\centering
\caption{Final optimized hyperparameters for Isolation Forest and Autoencoder models.}
\label{tab:unsupervisedhparams}
\begin{tabular}{lcc}
\toprule
\textbf{Isolation Forest}& \textbf{Declaration-level} & \textbf{Operation-level} \\
\midrule
Number of estimators         & 50    & 50 \\
Max sample size& 0.90  & 0.90 \\
Max feature proportion       & 0.10  & 0.10 \\
Bootstrapping                & False & False \\
Contamination threshold (\(\theta\)) & 0.016 & 0.016 \\
\midrule
\textbf{Autoencoder} & \textbf{Declaration-level} & \textbf{Operation-level} \\
\midrule
Optimizer                   & Adam     & RMSprop \\
Learning rate               & 0.01     & 0.01 \\
Hidden layer 1 neurons      & 32       & 16 \\
Hidden layer 2 neurons      & 10       & 8 \\
Activation (hidden layers)  & ReLU      & ReLU \\
Activation (output layer)   & Sigmoid   & Sigmoid \\
Early stopping patience     & 8 epochs & 8 epochs \\
Total training epochs& 28       & 18 \\
Anomaly threshold (percentile) & 98.4\% & 98.4\% \\
\bottomrule
\end{tabular}
\end{table}

\begin{table}[h]
\centering
\begin{minipage}[t]{0.48\textwidth}
\centering
\caption{Hyperparameter grid for Isolation Forest.}
\label{tab:iforesthyperparam}
\begin{tabular}{lc}
\toprule
\textbf{Hyperparameter} & \textbf{Value} \\
\midrule
Number of estimators & [50, 100, 150, 200, 300] \\
Max sample size & ['auto', 0.1, 0.3, 0.5, 0.7, 0.9] \\
Max feature proportion & [0.1, 0.3, 0.5, 0.7, 1.0] \\
Bootstrapping & [True, False] \\
\bottomrule
\end{tabular}
\end{minipage}
\hfill
\begin{minipage}[t]{0.48\textwidth}
\centering
\caption{Hyperparameter grid for Autoencoder.}
\label{tab:autoencoderhyperparam}
\begin{tabular}{lc}
\toprule
\textbf{Hyperparameter} & \textbf{Value} \\
\midrule
Hidden layer 1 neurons & [16, 32] \\
Hidden layer 2 neurons & [8, 10] \\
Learning rate & [0.001, 0.01, 0.005, 0.05] \\
Optimizer & ['adam', 'rmsprop', 'sgd'] \\
\bottomrule
\end{tabular}
\end{minipage}
\end{table}

\begin{table}[h]
\centering
\begin{minipage}[t]{0.48\textwidth}
\centering
\caption{Randomized search settings for Isolation Forest.}
\label{tab:iforestrandomsearch}
\begin{tabular}{lc}
\toprule
\textbf{Hyperparameter} & \textbf{Value} \\
\midrule
Number of iterations & 10 \\
Scoring & mean absolute deviation \\
cv & 3-fold \\
Verbose & 2 \\
Random state & 42 \\
\bottomrule
\end{tabular}
\end{minipage}
\hfill
\begin{minipage}[t]{0.48\textwidth}
\centering
\caption{Randomized search settings for Autoencoder.}
\label{tab:autoencoderrandomsearchDbased}
\begin{tabular}{lc}
\toprule
\textbf{Hyperparameter} & \textbf{Value} \\
\midrule
Epochs & 50 \\
Batch size & 32 \\
Callbacks & early stopping \\
Verbose & 1 \\
\bottomrule
\end{tabular}
\end{minipage}
\end{table}

\begin{table}[h]
\centering
\caption{LSTM hyperparameters and training outcomes for the declaration-level dataset.}
\label{tab:lstmhyperparam}
\begin{tabular}{lccc}
\toprule
\textbf{Hyperparameter}& \textbf{Autoencoder}& \textbf{Isolation Forest} &\textbf{Original-label}\\
\midrule
Window size& 100& 100&100\\
Batch size& 64& 64&64\\
Positive weight& 30& 30&10\\
Weight decay& 0.001& 0.001&0.001\\
Learning rate& 0.005& 0.005&0.005\\
Patience& 7& 7&7\\
Epochs run& 13& 50&19\\
Optimal used threshold& 0.875& 0.936&0.0003\\
Validation loss& 0.250& 0.170&1.976\\
Test loss& 0.331& 0.181&0.0003\\
\bottomrule
\end{tabular}
\end{table}

\begin{table}[h]
\centering
\caption{LSTM hyperparameters and training outcomes for the operation-level dataset. }
\label{tab:lstmhyperparamoperation}
\begin{tabular}{lccc}
\toprule
\textbf{Hyperparameter}& \textbf{Autoencoder}& \textbf{Isolation Forest} &\textbf{Original-label}\\
\midrule
Window size& 50& 50&100\\
Batch size& 64& 64&64\\
Positive weight& 30& 30&10\\
Weight decay& 0.001& 0.001&0.001\\
Learning rate& 0.0005& 0.0005&0.0005\\
Patience& 7& 7&7\\
Epochs run& 17& 43&8\\
Optimal used threshold& 0.296& 0.901&0.002\\
Validation loss& 0.057& 0.042&0.079\\
Test loss& 0.202& 0.063&0.031\\
\bottomrule
\end{tabular}
\end{table}

\begin{table}[h]
\centering
\caption{Transformer hyperparameters and training outcomes for the declaration-level dataset. }
\label{tab:transformerhyperparameter}
\begin{tabular}{lccc}
\toprule
\textbf{Hyperparameter}& \textbf{Autoencoder}& \textbf{Isolation Forest} &\textbf{Original-label}\\
\midrule
Window size& 100& 100&100\\
Batch size& 64& 64&64\\
Positive weight& 40& 40&20\\
Activation function& Gelu& Gelu&Gelu\\
Learning rate& 0.0001& 0.0001&0.0001\\
Patience& 5& 5&10\\
Epochs run& 14& 11&11\\
Optimal used threshold& 0.989& 0.990&0.000\\
Validation loss& 0.350& 0.062&6.402\\
Test loss& 0.317& 0.060&0.000\\
\bottomrule
\end{tabular}
\end{table}

\begin{table}[h]
\centering
\caption{Transformer hyperparameters and training outcomes for the operation-level dataset. }
\label{tab:transformerhyperparameteroperation}
\begin{tabular}{lccc}
\toprule
\textbf{Hyperparameter}& \textbf{Autoencoder}& \textbf{Isolation Forest} &\textbf{Original-label}\\
\midrule
Window size& 50& 50&100\\
Batch size& 64& 64&64\\
Positive weight& 20& 20&10\\
Activation function& Gelu& Gelu&Gelu\\
Learning rate& 0.0001& 0.0001&0.0005\\
Patience& 10& 10&10\\
Epochs run& 21& 25&18\\
Optimal used threshold& 0.749& 0.894&0.000\\
Validation loss& 0.059& 0.139&0.036\\
Test loss& 0.180& 0.362&0.006\\
\bottomrule
\end{tabular}
\end{table}

\section{Unsupervised anomaly detection}
\label{sec:appUnsupervisedanomalydetection}

\begin{table}[H]
\centering
\caption{Pseudo-label agreement between Isolation Forest and Autoencoder models for the declaration-level and operation-level datasets. Label 0 indicates normal instances, and label 1 indicates anomalous instances.}
\small 
\begin{tabular}{lrr}
\toprule
\textbf{Label combination}&\textbf{Count} & \textbf{Percentage}\\
\midrule
\multicolumn{3}{l}{\textbf{Declaration-level dataset}} \\
Both label 1 & 184& 0.26\%\\
Both label 0& 69,100&97.06\% \\
iForest 1 and AE 0 &956&1.34\%  \\
iForest 0 and AE 1&956&1.34\%\\
\textbf{Total} &\textbf{71,196} & \textbf{100\%}\\
\midrule
\multicolumn{3}{l}{\textbf{Operation-level dataset}} \\
Both label 1 & 4,581& 0.29\%\\
Both label 0& 1,543,664&97.09\% \\
iForest 1 and AE 0 &20,859&1.31\%  \\
iForest 0 and AE 1&20,847&1.31\%\\
\textbf{Total} &\textbf{1,589,951} & \textbf{100\%}\\
\bottomrule
\end{tabular}
\label{tab:pseudolabeldiff}
\end{table}

\begin{table*}[ht]
\centering
\caption{Summary statistics for selected features among the top 10 anomalies for the declaration-level dataset.}
\label{tab:topanomalystatsdeclaration}
\small
\begin{tabular}{llrrrr}
\toprule
\textbf{Model} & \textbf{Feature} & \textbf{Mean} & \textbf{Std} & \textbf{Min} & \textbf{Max} \\
\midrule
\multirow{6}{*}{Isolation Forest} 
  & Anomaly score & $-8.80\text{e}\text{-}2$ & $6.35\text{e}\text{-}3$ & $-1.02\text{e}\text{-}1$ & $-8.13\text{e}\text{-}2$ \\
  & AAccep        & $2.97\text{e}8$          & $3.93\text{e}8$        & $0.00$            & $9.00\text{e}8$ \\
  & ASup          & $3.01\text{e}8$          & $3.91\text{e}8$        & $6.20\text{e}6$          & $9.02\text{e}8$ \\
  & TOAccep       & $3.11\text{e}6$          & $4.32\text{e}6$        & $0.00$            & $9.75\text{e}6$ \\
  & TOStop        & $3.50\text{e}4$          & $3.86\text{e}4$        & $1.70\text{e}3$          & $1.28\text{e}5$ \\
  & AInsu         & $1.79\text{e}4$          & $3.72\text{e}4$        & $0.00$            & $1.06\text{e}5$ \\
\midrule
\multirow{6}{*}{Autoencoder} 
  & Recon error   & $2.21\text{e}\text{-}1$ & $5.15\text{e}\text{-}2$ & $1.50\text{e}\text{-}1$ & $3.40\text{e}\text{-}1$ \\
  & AAccep        & $6.50\text{e}8$ & $6.40\text{e}8$ & $0.00$ & $1.88\text{e}9$ \\
  & ASup          & $6.79\text{e}8$ & $6.05\text{e}8$ & $2.72\text{e}3$ & $1.88\text{e}9$ \\
  & TOAccep       & $6.25\text{e}6$ & $6.13\text{e}6$ & $0.00$ & $1.84\text{e}7$ \\
  & TOStop        & $4.92\text{e}5$ & $7.88\text{e}5$ & $0.00$ & $1.64\text{e}6$ \\
  & AInsu         & $0.00$ & $0.00$ & $0.00$ & $0.00$ \\
\bottomrule
\end{tabular}
\end{table*}

\begin{table*}[ht]
\centering
\caption{Summary statistics for selected features among the top 10 anomalies for the operation-level dataset.}
\label{tab:topanomalystatsoperation}
\small
\begin{tabular}{llrrrr}
\toprule
\textbf{Model} & \textbf{Feature} & \textbf{Mean} & \textbf{Std} & \textbf{Min} & \textbf{Max} \\
\midrule
\multirow{6}{*}{Isolation Forest} 
  & Anomaly score & $-6.25\text{e}\text{-}2$ & $8.04\text{e}\text{-}4$ & $-6.35\text{e}\text{-}2$ & $-6.13\text{e}\text{-}2$ \\
  & AAccep        & $0.00$                  & $0.00$                  & $0.00$                  & $0.00$ \\
  & ASup          & $1.85\text{e}4$         & $1.53\text{e}4$         & $5.00$                  & $3.03\text{e}4$ \\
  & TOAccep       & $0.00$                  & $0.00$                  & $0.00$                  & $0.00$ \\
  & TOStop        & $1.07\text{e}2$         & $8.75\text{e}1$         & $1.00$                  & $1.75\text{e}2$ \\
  & AInsu         & $0.00$                  & $0.00$                  & $0.00$                  & $0.00$ \\
\midrule
\multirow{6}{*}{Autoencoder} 
  & Recon error   & $1.13\text{e}\text{-}1$ & $2.37\text{e}\text{-}3$ & $1.10\text{e}\text{-}1$ & $1.16\text{e}\text{-}1$ \\
  & AAccep        & $8.33\text{e}6$         & $6.45\text{e}6$         & $0.00$                  & $1.41\text{e}7$ \\
  & ASup          & $8.60\text{e}6$         & $6.67\text{e}6$         & $3.17\text{e}3$         & $1.44\text{e}7$ \\
  & TOAccep       & $5.26\text{e}4$         & $4.17\text{e}4$         & $0.00$                  & $9.45\text{e}4$ \\
  & TOStop        & $2.37\text{e}3$         & $2.10\text{e}3$         & $1.60\text{e}1$         & $4.44\text{e}3$ \\
  & AInsu         & $7.33\text{e}4$         & $6.33\text{e}4$         & $0.00$                  & $1.27\text{e}5$ \\
\bottomrule
\end{tabular}
\end{table*}

\begin{figure}[h!]
\centering
\begin{minipage}[t]{0.48\textwidth}
  \centering
  \includegraphics[width=0.8\linewidth, height=8cm, keepaspectratio]{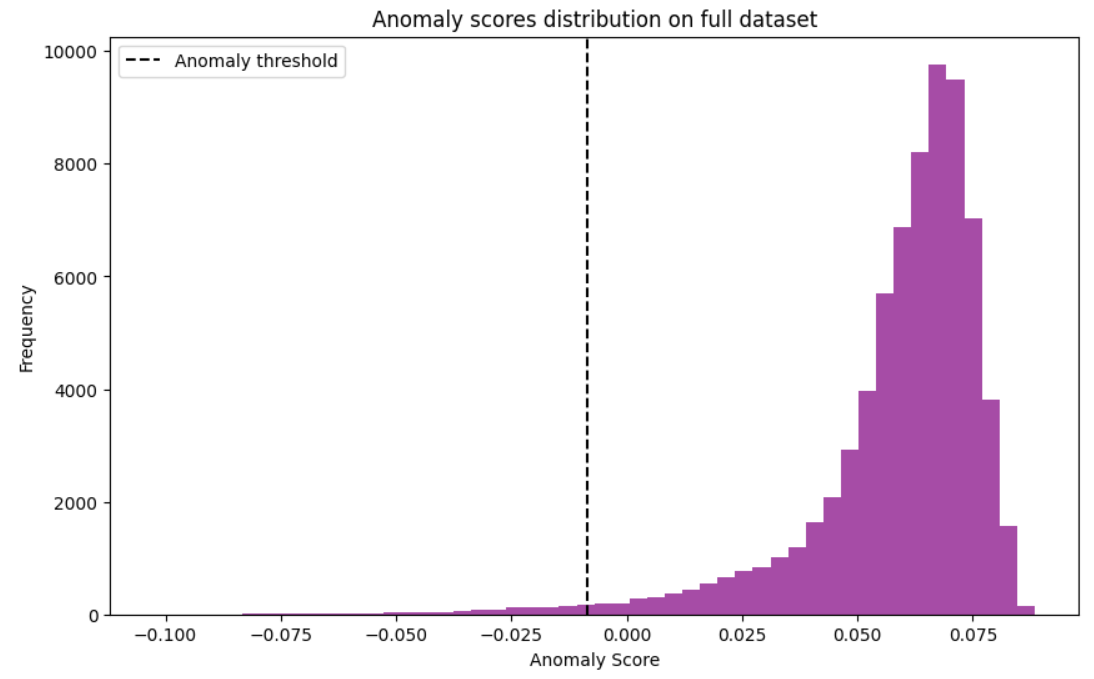}
  \caption{Anomaly score distribution of Isolation Forest for the declaration-level dataset.}
  \label{fig:iforestplot}
\end{minipage}
\hfill
\begin{minipage}[t]{0.48\textwidth}
  \centering
  \includegraphics[width=0.8\linewidth, height=8cm, keepaspectratio]{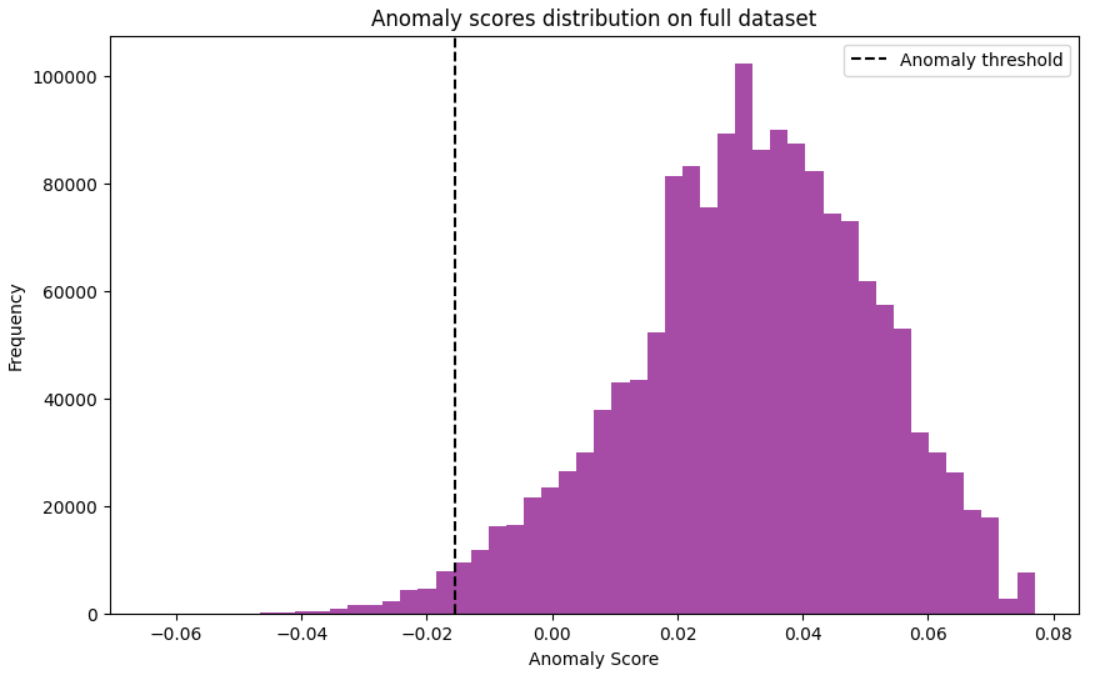}
  \caption{Anomaly score distribution of Isolation Forest for the operation-level dataset.}
  \label{fig:iforestplotobased}
\end{minipage}
\end{figure}

\begin{figure}[h!]
\centering
\begin{minipage}[t]{0.48\textwidth}
  \centering
  \includegraphics[width=0.8\linewidth, height=8cm, keepaspectratio]{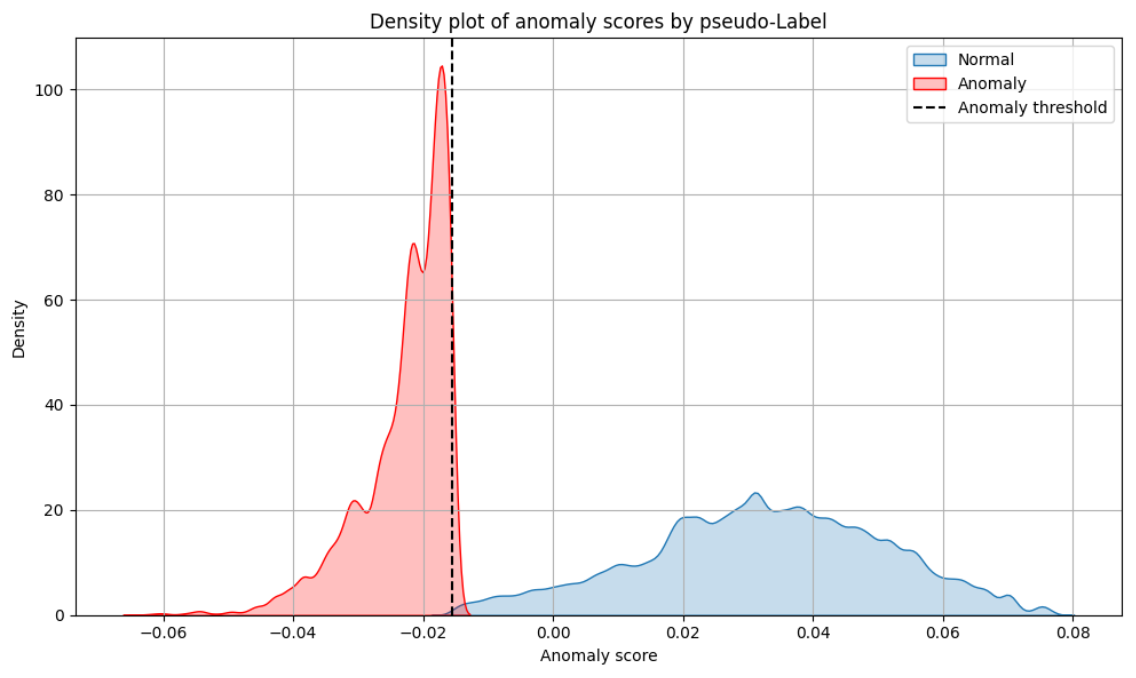}
  \caption{Density plot of Isolation Forest anomaly scores for the operation-level dataset. }
  \label{fig:iforestdensityobased}
\end{minipage}
\hfill
\begin{minipage}[t]{0.48\textwidth}
  \centering
  \includegraphics[width=0.8\linewidth, height=8cm, keepaspectratio]{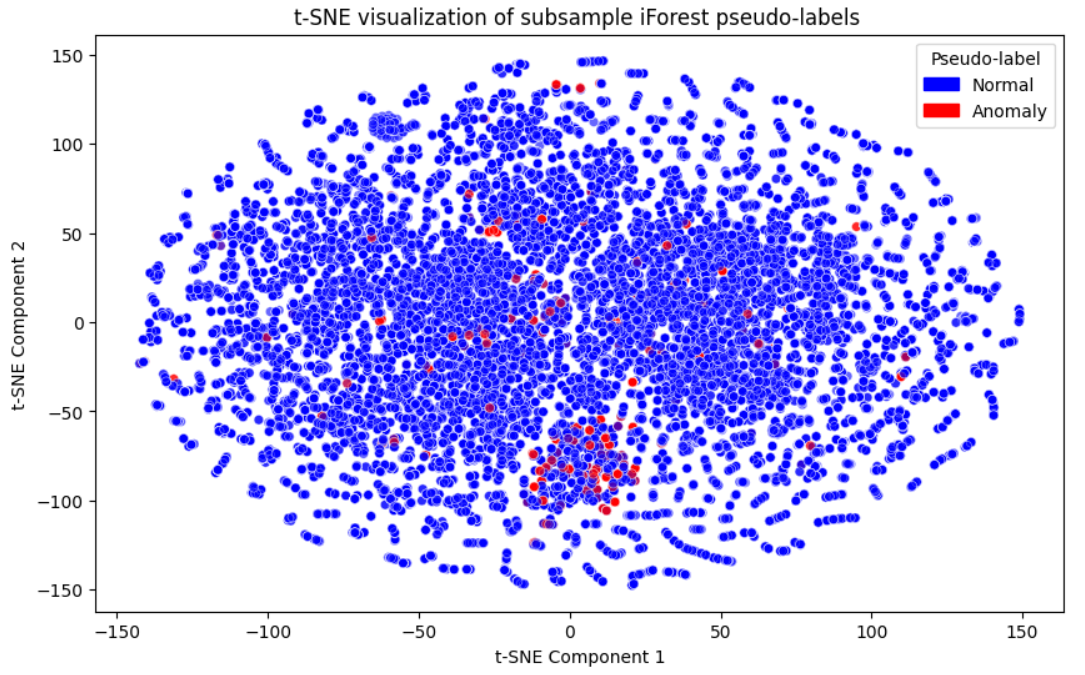}
  \caption{T-SNE plot of Isolation Forest pseudo-labels for the operation-level dataset.}
  \label{fig:iforesttsneobased}
\end{minipage}
\end{figure}

\begin{figure}[h!]
\centering
\begin{minipage}[t]{0.48\textwidth}
  \centering
  \includegraphics[width=0.8\linewidth, height=8cm, keepaspectratio]{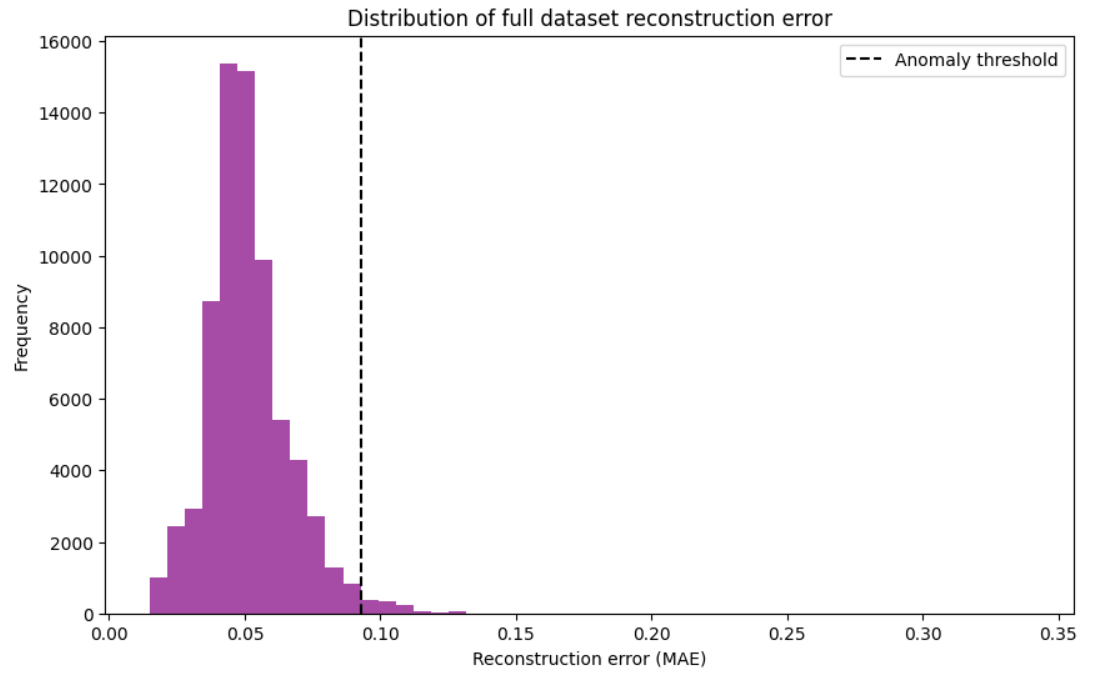}
  \caption{Reconstruction error distribution of Autoencoder for the declaration-level dataset.}
  \label{fig:autoencoderreconerror}
\end{minipage}
\hfill
\begin{minipage}[t]{0.48\textwidth}
  \centering
  \includegraphics[width=0.8\linewidth, height=8cm, keepaspectratio]{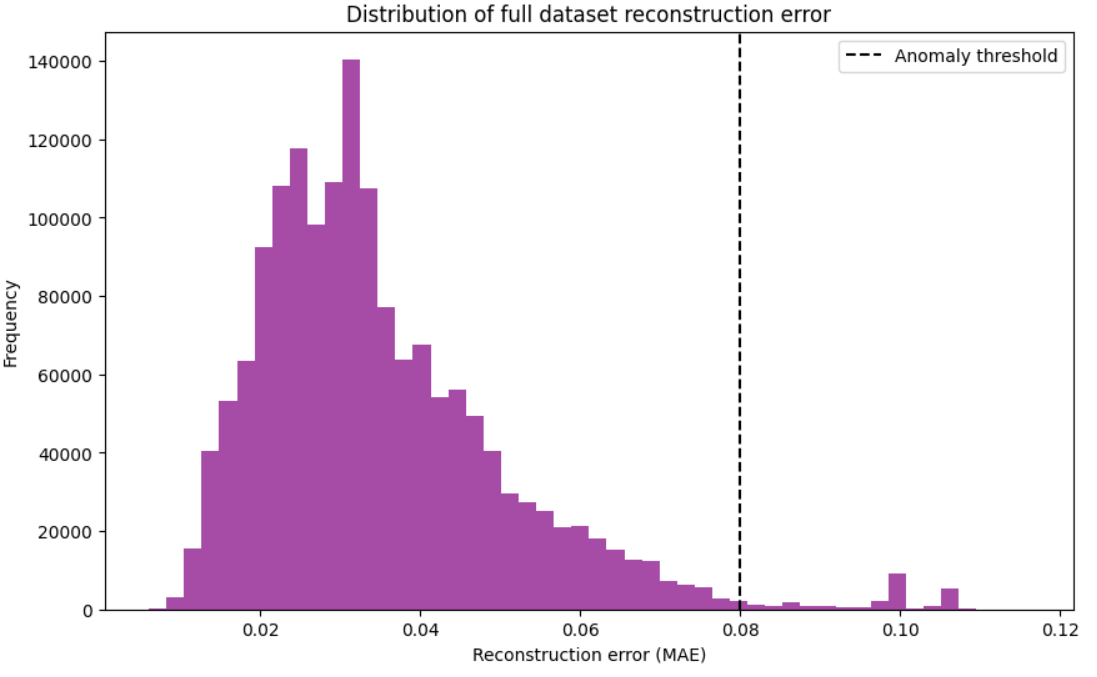}
  \caption{Reconstruction error distribution of Autoencoder for the operation-level dataset.}
  \label{fig:autoencoderreconerrorobased}
\end{minipage}
\end{figure}

\begin{figure}[h!]
\centering
\begin{minipage}[t]{0.48\textwidth}
  \centering
  \includegraphics[width=0.8\linewidth, height=8cm, keepaspectratio]{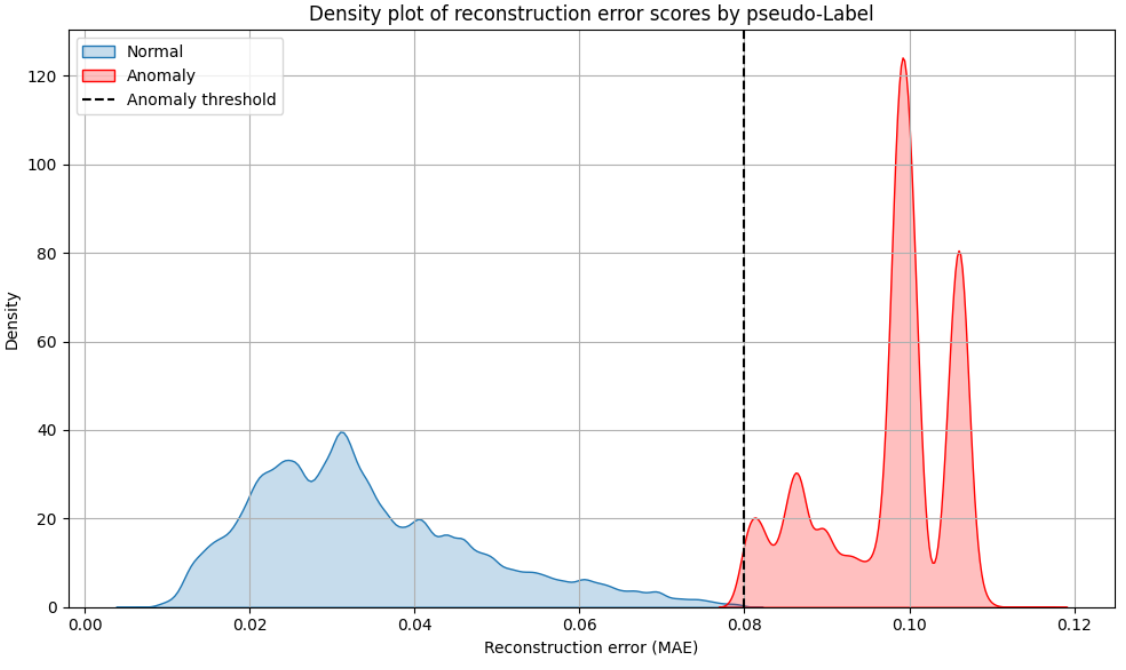}
  \caption{Density plot of Autoencoder reconstruction errors for the operation-level dataset.}
  \label{fig:autoencoderdensityobased}
\end{minipage}
\hfill
\begin{minipage}[t]{0.48\textwidth}
  \centering
  \includegraphics[width=0.8\linewidth, height=8cm, keepaspectratio]{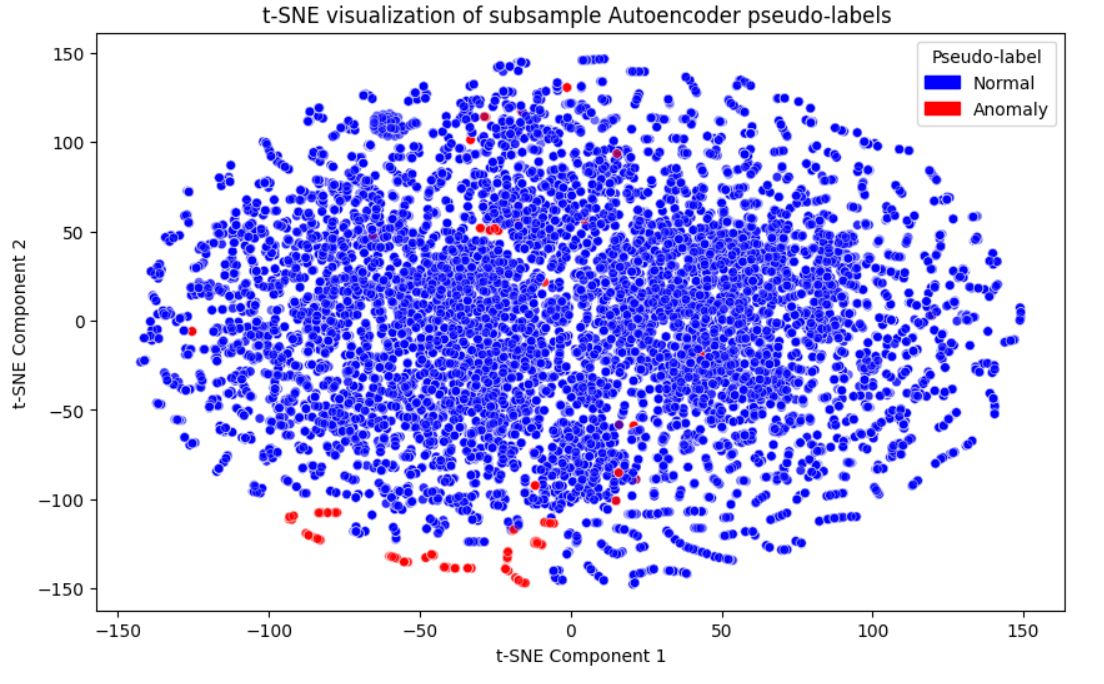}
  \caption{T-SNE plot of Autoencoder pseudo-labels for the operation-level dataset.}
  \label{fig:autoencodertsneobased}
\end{minipage}
\end{figure}

\section{Model performance}
\label{sec:evalmetricsappendix}

\begin{table}[H]
\centering
\begin{minipage}{0.48\textwidth}
\centering
\caption{McNemar's test results for detecting significant differences in classification errors between LSTM and Transformer models on the \textbf{test set}.}
\small
\begin{tabular}{lccc}
\toprule
\textbf{Pseudo-label model} & \textbf{Statistic} & \textbf{P-value} & \textbf{Significant} \\
\midrule
\multicolumn{4}{l}{\textbf{Declaration-level dataset}} \\
Autoencoder & 21.25 & 0.00 & Yes \\
Isolation Forest & 57.14 & 0.00 & Yes \\
\midrule
\multicolumn{4}{l}{\textbf{Operation-level dataset}} \\
Autoencoder & 38.40 & 0.00 & Yes \\
Isolation Forest & 41.40 & 0.00 & Yes \\
\bottomrule
\end{tabular}
\label{tab:mcnemar_test}
\end{minipage}
\hfill
\begin{minipage}{0.48\textwidth}
\centering
\caption{McNemar's test results for detecting significant differences in classification errors between LSTM and Transformer models on the \textbf{validation set}.}
\small
\begin{tabular}{lccc}
\toprule
\textbf{Pseudo-label model} & \textbf{Statistic} & \textbf{P-value} & \textbf{Significant} \\
\midrule
\multicolumn{4}{l}{\textbf{Declaration-level dataset}} \\
Autoencoder & 38.40 & 0.00 & Yes \\
Isolation Forest & 41.40 & 0.00 & Yes \\
\midrule
\multicolumn{4}{l}{\textbf{Operation-level dataset}} \\
Autoencoder & 121.18 & 0.00 & Yes \\
Isolation Forest & 5187.17 & 0.00 & Yes \\
\bottomrule
\end{tabular}
\label{tab:mcnemar_val}
\end{minipage}
\end{table}

\begin{table}[H]
\centering
\caption{Overview of model performance on the declaration-level dataset, based on evaluation metrics computed on the validation set.}
\small 
\begin{tabular}{lcccc}
\toprule
\textbf{Classifiers} & \textbf{Accuracy} & \textbf{Precision} & \textbf{Recall} & \textbf{F1-score} \\
\midrule
\multicolumn{5}{l}{\textbf{Baseline models}} \\
AE Transformer & 0.993& 0.802& 0.478& 0.599\\
iForest Transformer& 0.995 & 0.868& 0.865& 0.867\\
Original-label Transformer& 0.995& 0.219& 0.315& 0.258\\
AE LSTM & 0.988 & 0.462& 0.565& 0.508\\
iForest LSTM& \textbf{0.999}& \textbf{0.960}& \textbf{0.980}& \textbf{0.970}\\
Original-label LSTM& 0.634& 0.043& 0.656& 0.081\\
\bottomrule
\end{tabular}
\label{tab:evaldbasedval}
\end{table}

\begin{table}[ht]
\centering
\caption{Overview of model performance on the operation-level dataset, based on evaluation metrics computed on the validation set.}
\small 
\begin{tabular}{lcccc}
\toprule
\textbf{Classifiers} & \textbf{Accuracy} & \textbf{Precision} & \textbf{Recall} & \textbf{F1-score} \\
\midrule
\multicolumn{5}{l}{\textbf{Baseline models}} \\
AE Transformer & 0.998& 0.348& 0.846& 0.494\\
iForest Transformer& 0.984& 0.396& 0.661& 0.495\\
Original-label Transformer& \textbf{0.999}& 0.000& 0.000& 0.000\\
AE LSTM & 0.998& 0.427& 0.867& 0.572\\
iForest LSTM & 0.997& \textbf{0.881}& \textbf{0.878}& \textbf{0.879}\\
Original-label LSTM& 0.997& 0.056& 0.295& 0.094\\
\bottomrule
\end{tabular}
\label{tab:evalobasedval}
\end{table}

\section{SHAP explanation}

\label{sec:shapappendix}

\begin{figure}[h!]
\centering
\begin{minipage}[t]{0.48\textwidth}
  \centering
  \includegraphics[width=0.8\linewidth, height=8cm, keepaspectratio]{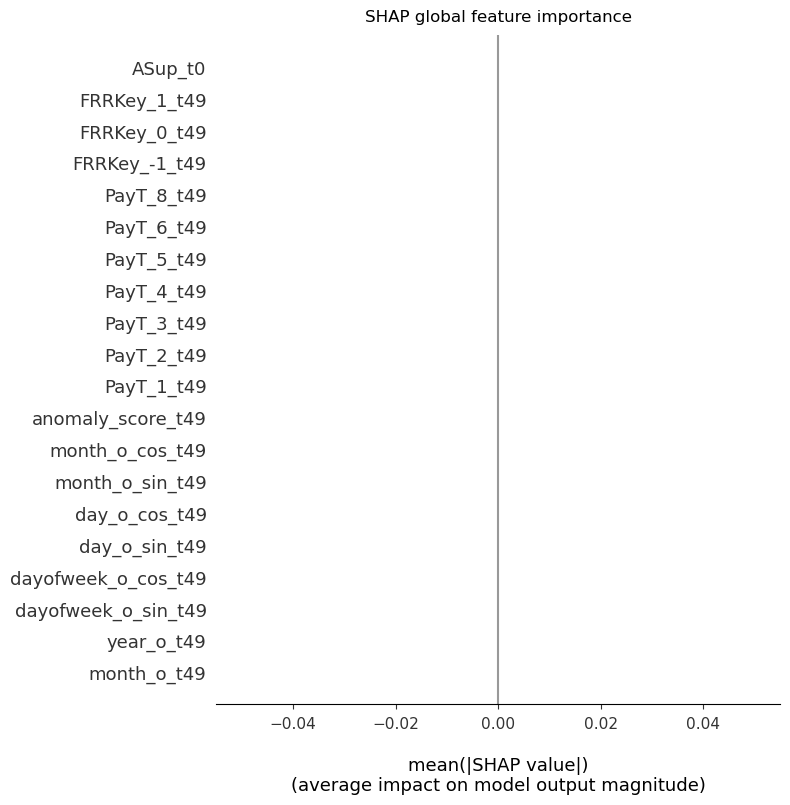}
  \caption{SHAP summary bar plot for Isolation Forest LSTM for the operation-level dataset.}
  \label{fig:shapiforestlstmobased}
\end{minipage}
\hfill
\begin{minipage}[t]{0.48\textwidth}
  \centering
  \includegraphics[width=0.8\linewidth, height=8cm, keepaspectratio]{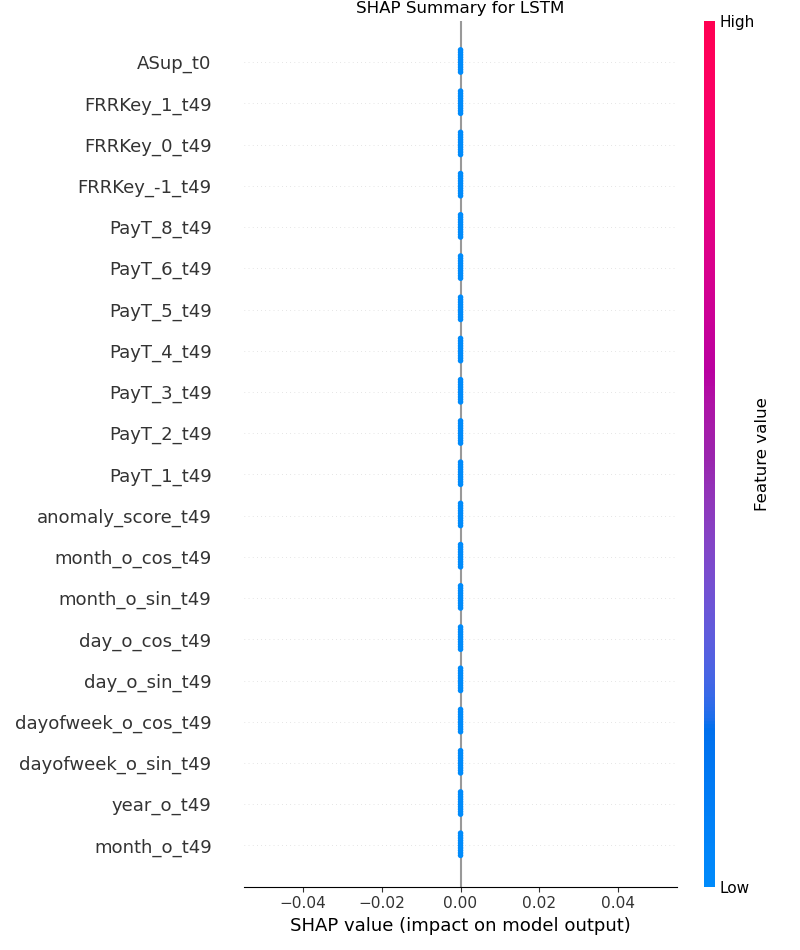}
  \caption{SHAP summary plot for Isolation Forest LSTM for the operation-level dataset.}
  \label{fig:shapsumiforestlstmobased}
\end{minipage}
\end{figure}

\begin{figure}[h!]
\centering
\begin{minipage}[t]{0.48\textwidth}
  \centering
  \includegraphics[width=0.8\linewidth, height=8cm, keepaspectratio]{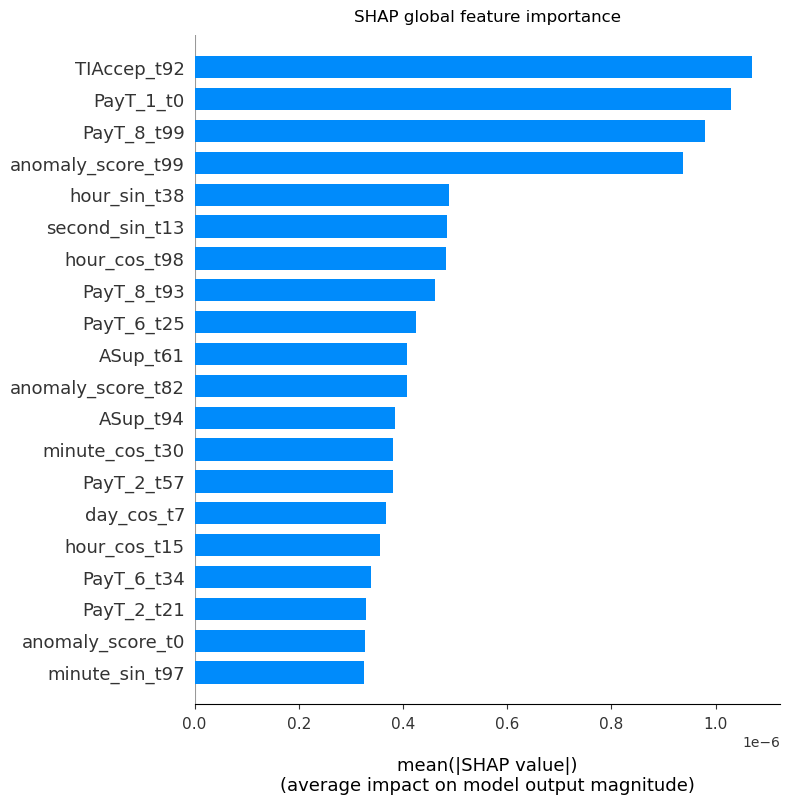}
  \caption{SHAP summary bar plot for Isolation Forest Transformer for the declaration-level dataset.}
  \label{fig:shapiforesttransbased}
\end{minipage}
\hfill
\begin{minipage}[t]{0.48\textwidth}
  \centering
  \includegraphics[width=0.8\linewidth, height=8cm, keepaspectratio]{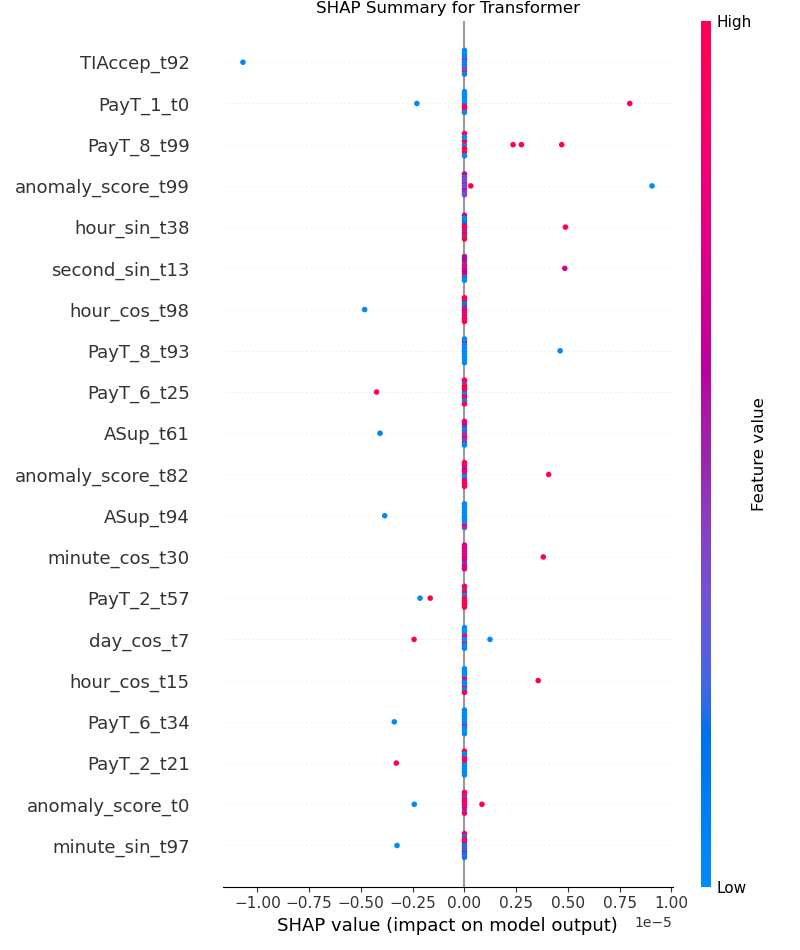}
  \caption{SHAP summary plot for Isolation Forest Transformer for the declaration-level dataset.}
  \label{fig:shapsumiforesttransbased}
\end{minipage}
\end{figure}

\begin{figure}[h!]
\centering
\begin{minipage}[t]{0.48\textwidth}
  \centering
  \includegraphics[width=0.8\linewidth, height=8cm, keepaspectratio]{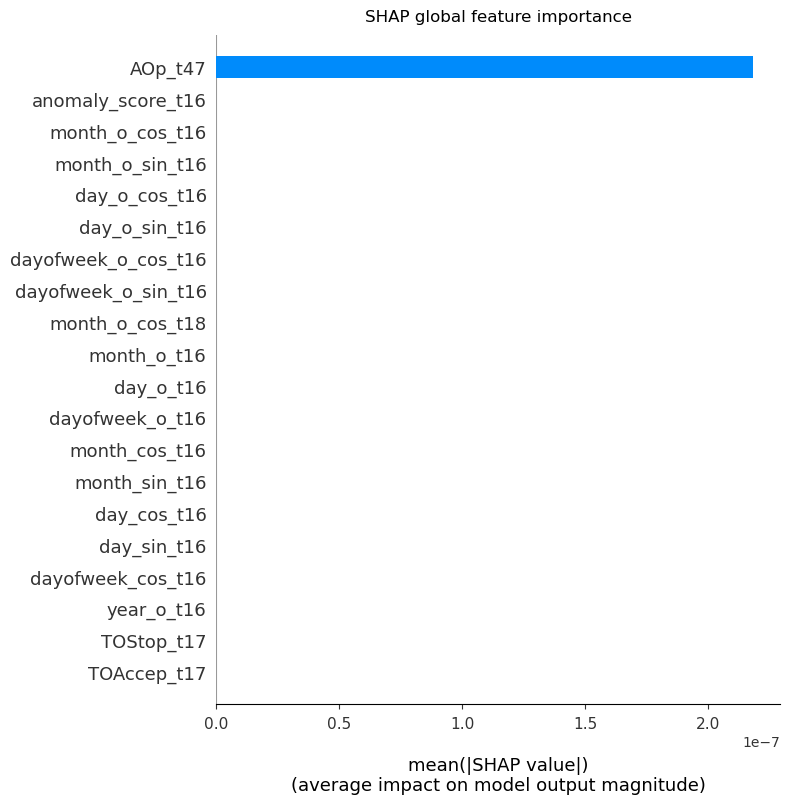}
  \caption{SHAP summary bar plot for Isolation Forest Transformer for the operation-level dataset.}
  \label{fig:shapiforesttransbasedo}
\end{minipage}
\hfill
\begin{minipage}[t]{0.48\textwidth}
  \centering
  \includegraphics[width=0.8\linewidth, height=8cm, keepaspectratio]{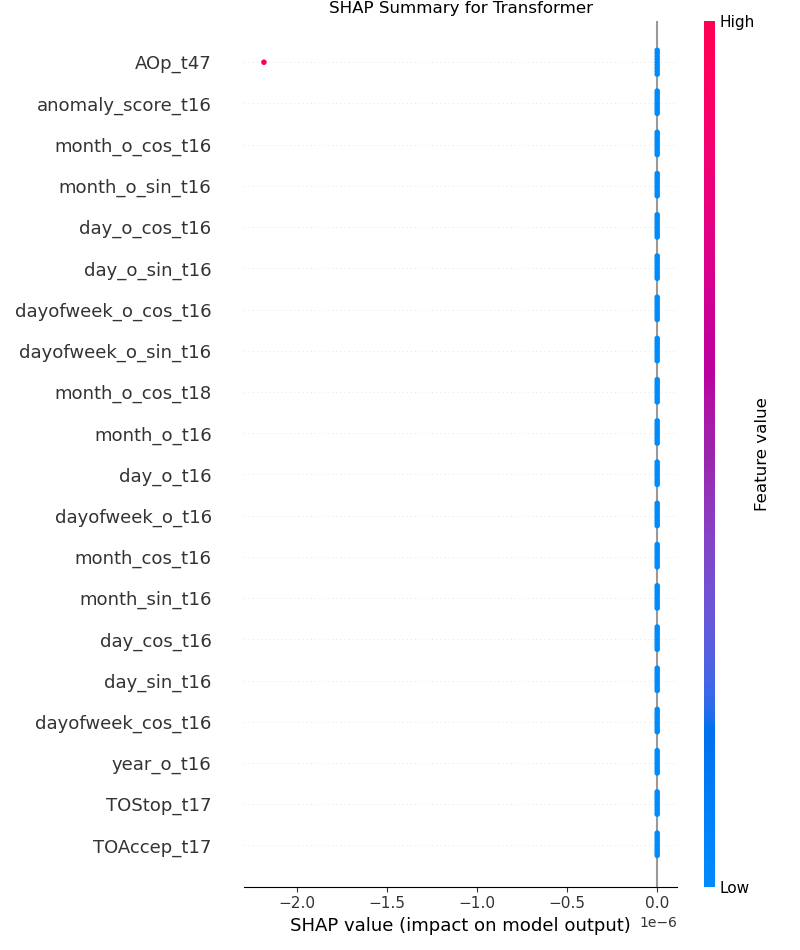}
  \caption{SHAP summary plot for Isolation Forest Transformer for the operation-level dataset.}
  \label{fig:shapsumiforesttransbasedo}
\end{minipage}
\end{figure}

\begin{figure}[h!]
\centering
\begin{minipage}[t]{0.48\textwidth}
  \centering
  \includegraphics[width=0.8\linewidth, height=8cm, keepaspectratio]{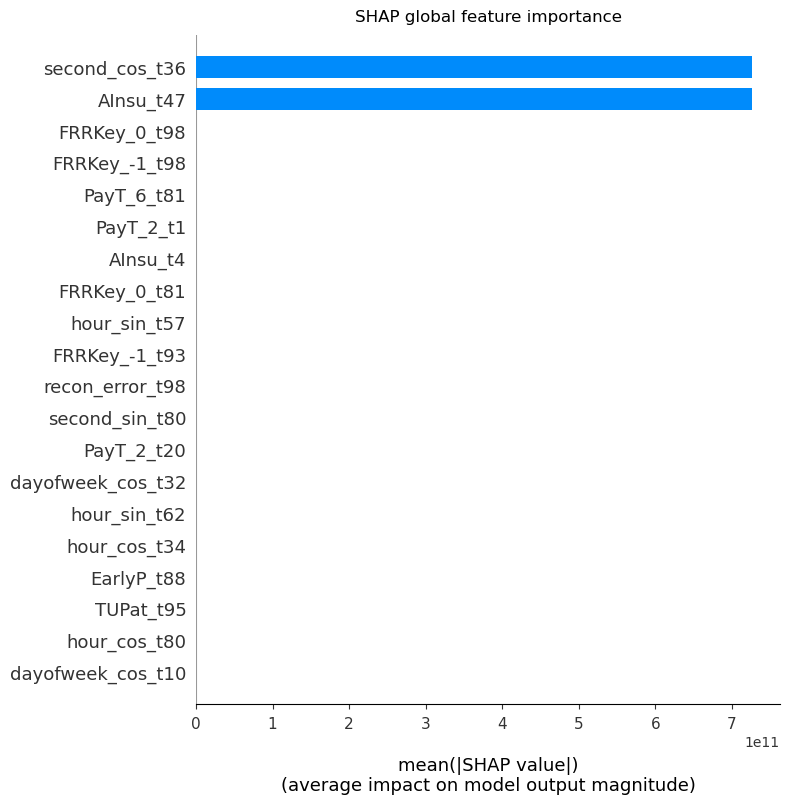}
  \caption{SHAP summary bar plot for Autoencoder LSTM for the declaration-level dataset.}
  \label{fig:shapaelstmbased}
\end{minipage}
\hfill
\begin{minipage}[t]{0.48\textwidth}
  \centering
  \includegraphics[width=0.8\linewidth, height=8cm, keepaspectratio]{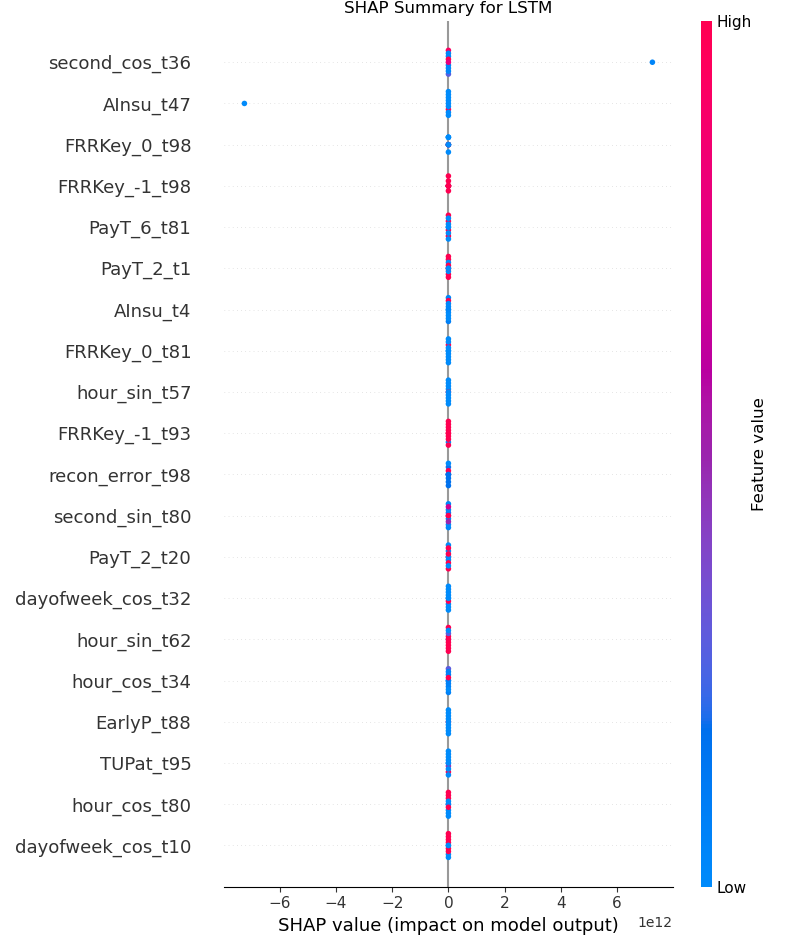}
  \caption{SHAP summary plot for Autoencoder LSTM for the declaration-level dataset.}
  \label{fig:shapsumaelstmbased}
\end{minipage}
\end{figure}

\begin{figure}[h!]
\centering
\begin{minipage}[t]{0.48\textwidth}
  \centering
  \includegraphics[width=0.8\linewidth, height=8cm, keepaspectratio]{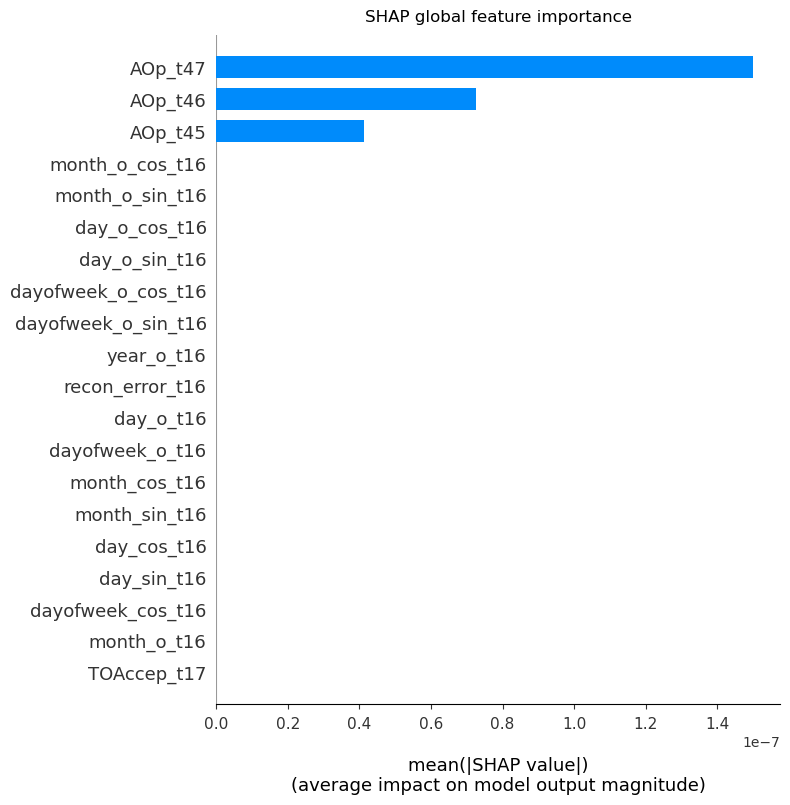}
  \caption{SHAP summary bar plot for Autoencoder LSTM for the operation-level dataset.}
  \label{fig:shapaelstmbasedo}
\end{minipage}
\hfill
\begin{minipage}[t]{0.48\textwidth}
  \centering
  \includegraphics[width=0.8\linewidth, height=8cm, keepaspectratio]{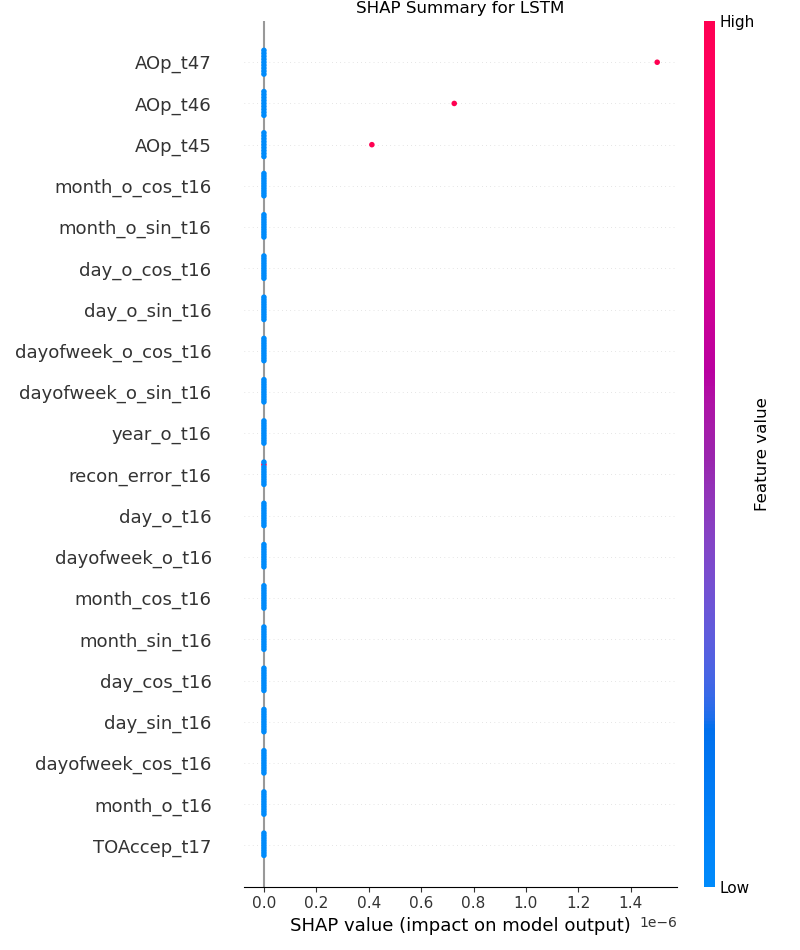}
  \caption{SHAP summary plot for Autoencoder LSTM for the operation-level dataset.}
  \label{fig:shapsumaelstmbasedo}
\end{minipage}
\end{figure}

\begin{figure}[h!]
\centering
\begin{minipage}[t]{0.48\textwidth}
  \centering
  \includegraphics[width=0.8\linewidth, height=8cm, keepaspectratio]{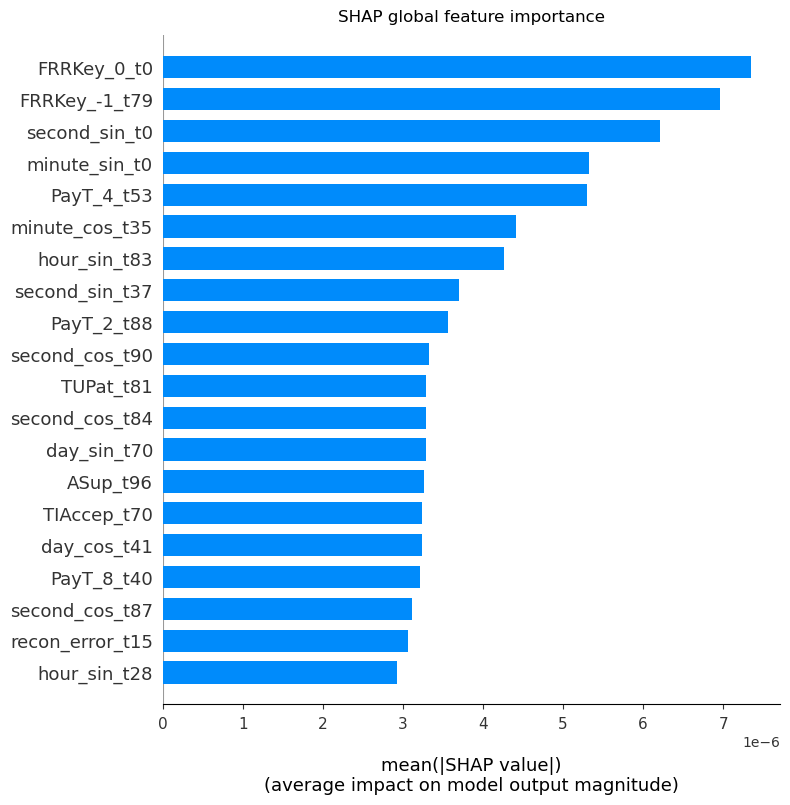}
  \caption{SHAP summary bar plot for Autoencoder Transformer for the declaration-level dataset.}
  \label{fig:shapaetransbased}
\end{minipage}
\hfill
\begin{minipage}[t]{0.48\textwidth}
  \centering
  \includegraphics[width=0.8\linewidth, height=8cm, keepaspectratio]{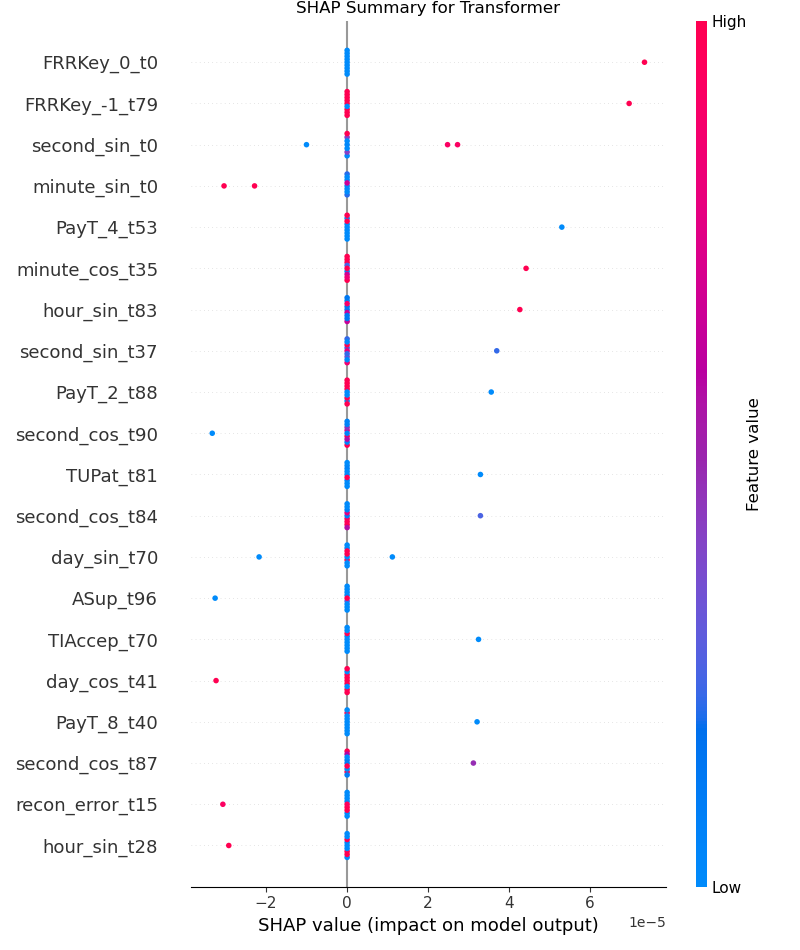}
  \caption{SHAP summary plot for Autoencoder Transformer for the declaration-level dataset.}
  \label{fig:shapsumaetransbased}
\end{minipage}
\end{figure}

\begin{figure}[h!]
\centering
\begin{minipage}[t]{0.48\textwidth}
  \centering
  \includegraphics[width=0.8\linewidth, height=8cm, keepaspectratio]{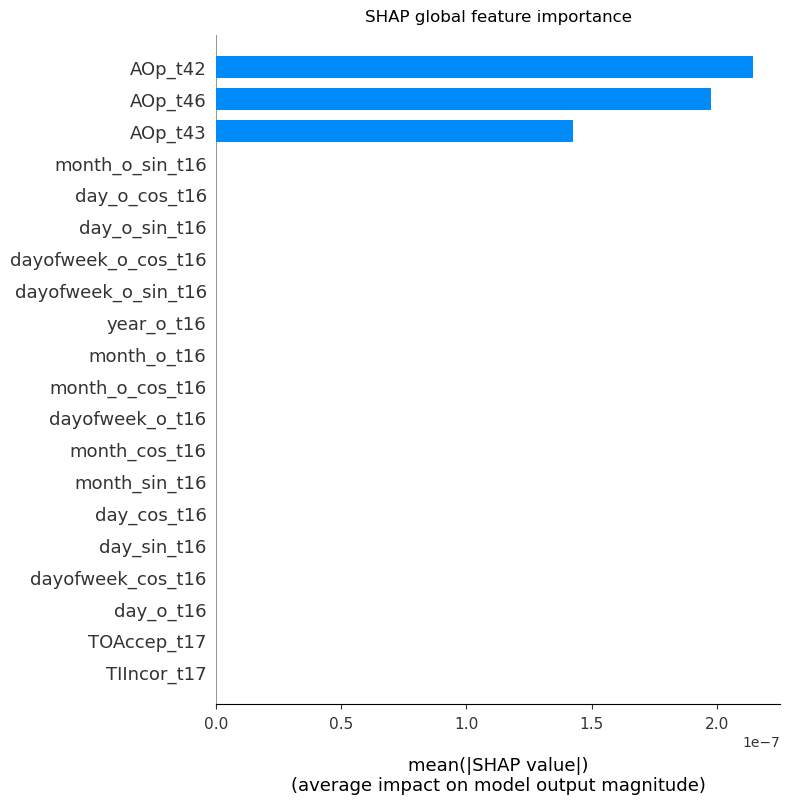}
  \caption{SHAP summary bar plot for Autoencoder Transformer for the operation-level dataset.}
  \label{fig:shapaetransbasedo}
\end{minipage}
\hfill
\begin{minipage}[t]{0.48\textwidth}
  \centering
  \includegraphics[width=0.8\linewidth, height=8cm, keepaspectratio]{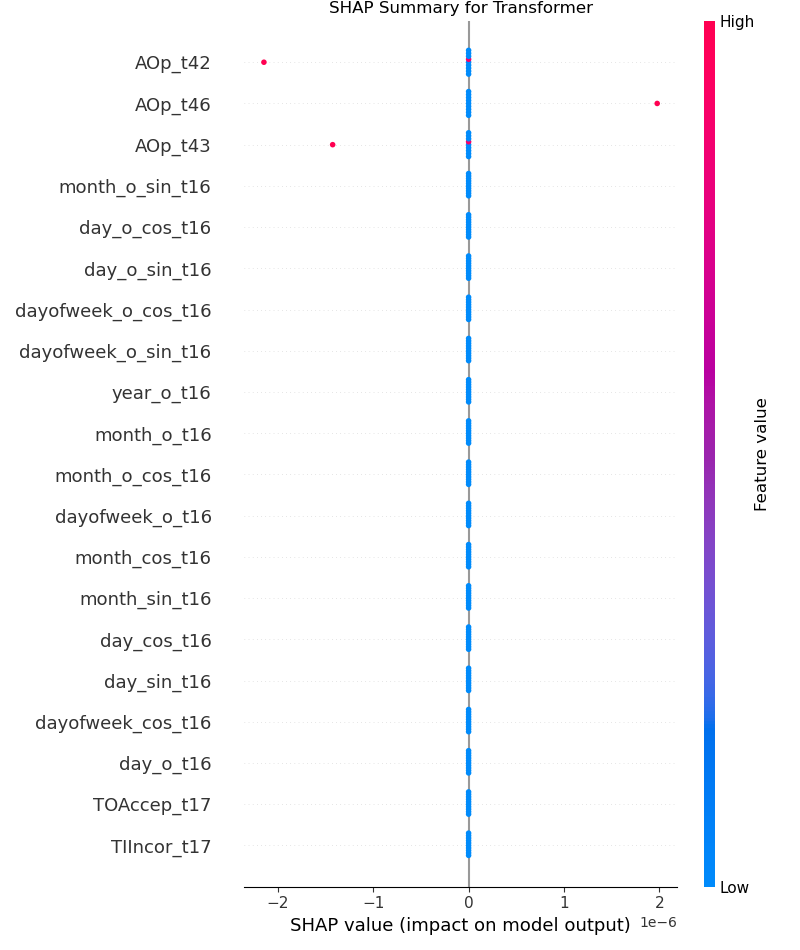}
  \caption{SHAP summary plot for Autoencoder Transformer for the operation-level dataset.}
  \label{fig:shapsumaetransbasedo}
\end{minipage}
\end{figure}

\begin{figure}[h!]
\centering
\begin{minipage}[t]{0.48\textwidth}
  \centering
  \includegraphics[width=0.8\linewidth, height=8cm, keepaspectratio]{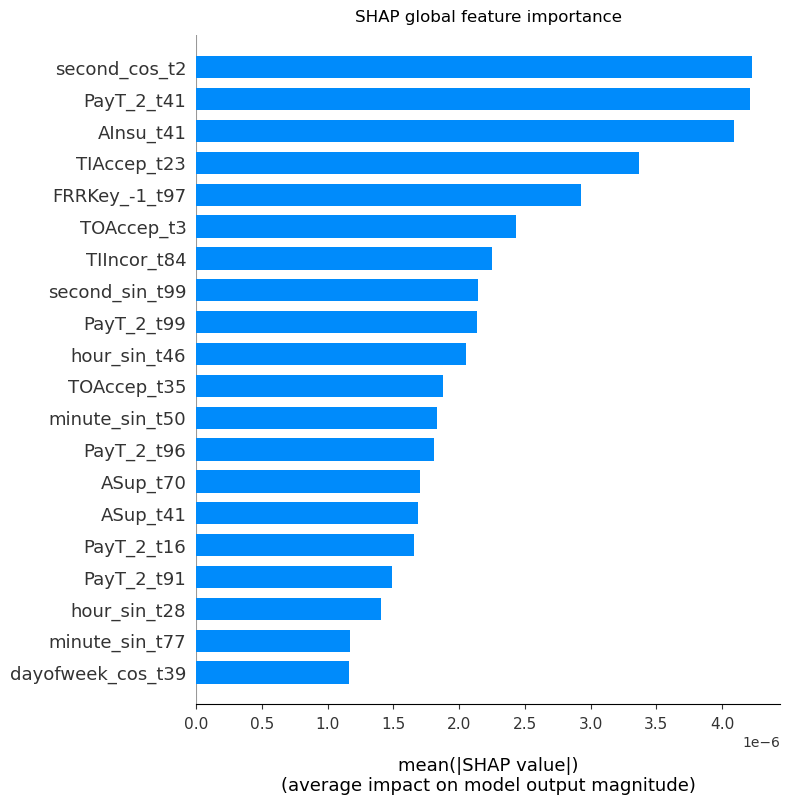}
  \caption{SHAP summary bar plot for original-label LSTM for the declaration-level dataset.}
  \label{fig:shaporiglstmbased}
\end{minipage}
\hfill
\begin{minipage}[t]{0.48\textwidth}
  \centering
  \includegraphics[width=0.8\linewidth, height=8cm, keepaspectratio]{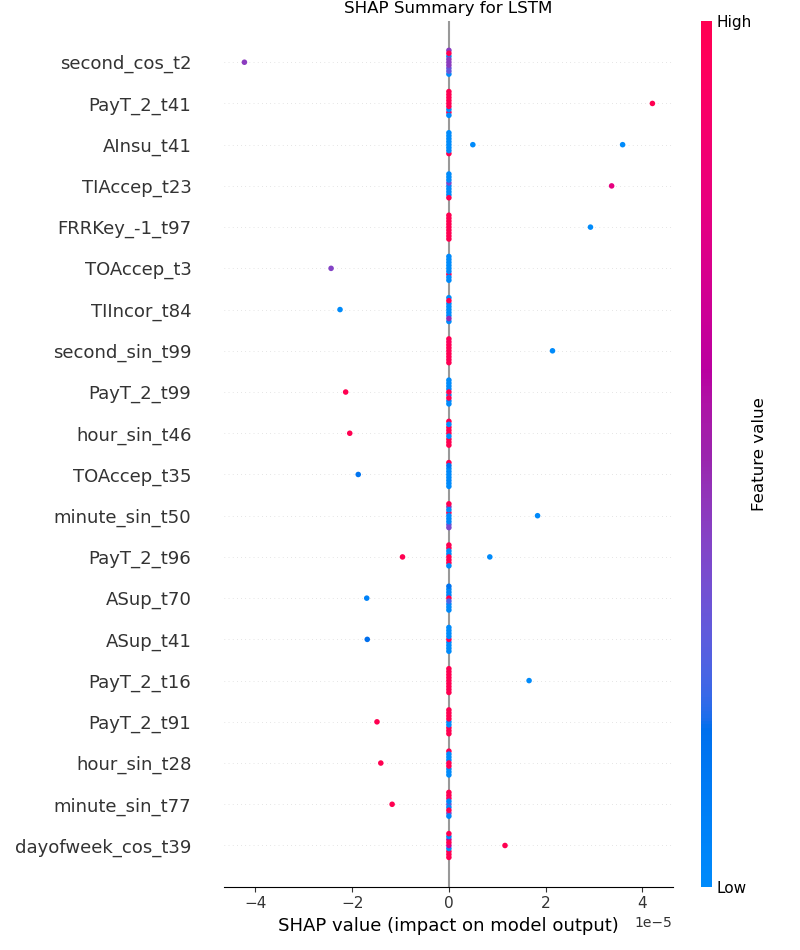}
  \caption{SHAP summary plot for original-label LSTM for the declaration-level dataset.}
  \label{fig:shapsumoriglstmbased}
\end{minipage}
\end{figure}

\begin{figure}[h!]
\centering
\begin{minipage}[t]{0.48\textwidth}
  \centering
  \includegraphics[width=0.8\linewidth, height=8cm, keepaspectratio]{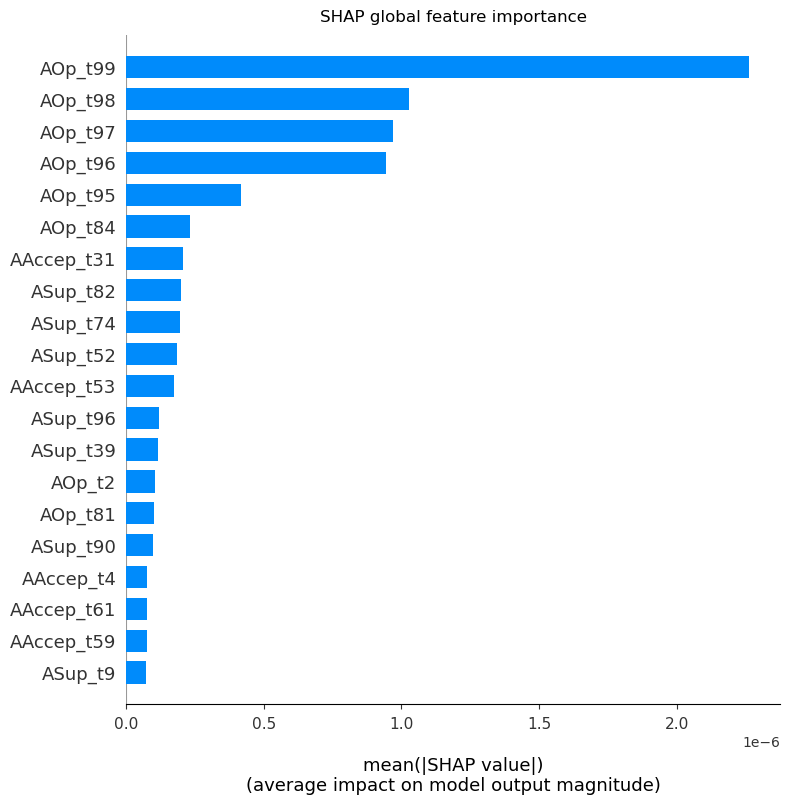}
  \caption{SHAP summary bar plot for original-label LSTM for the operation-level dataset.}
  \label{fig:shaporiglstmbasedo}
\end{minipage}
\hfill
\begin{minipage}[t]{0.48\textwidth}
  \centering
  \includegraphics[width=0.8\linewidth, height=8cm, keepaspectratio]{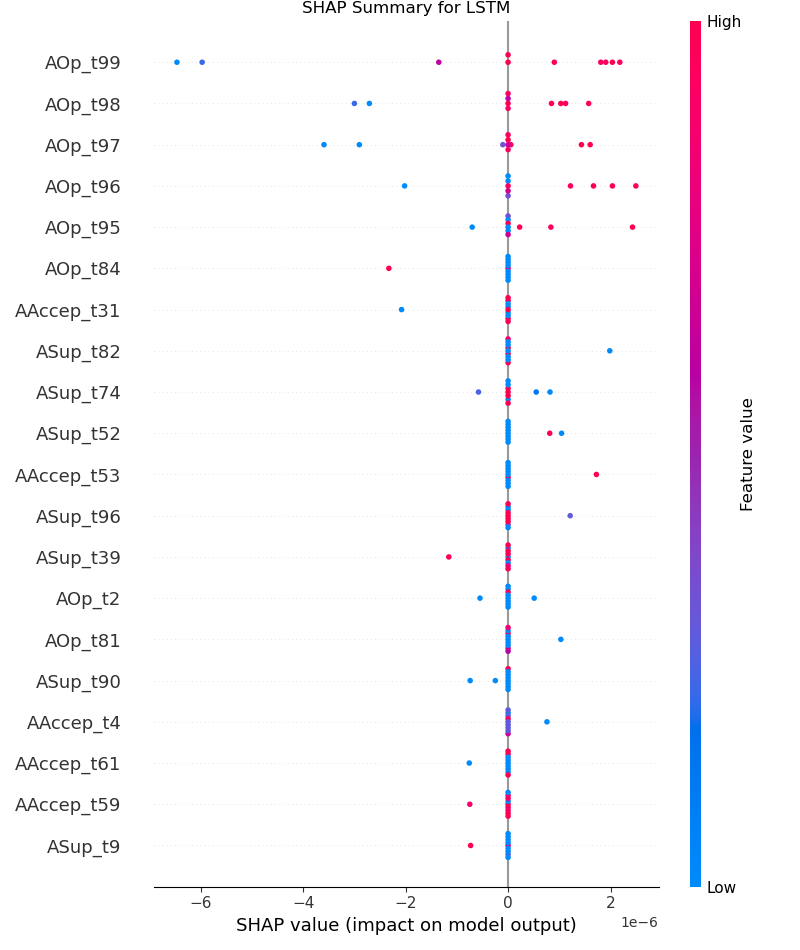}
  \caption{SHAP summary plot for original-label LSTM for the operation-level dataset.}
  \label{fig:shapsumoriglstmbasedo}
\end{minipage}
\end{figure}

\begin{figure}[h!]
\centering
\begin{minipage}[t]{0.48\textwidth}
  \centering
  \includegraphics[width=0.8\linewidth, height=8cm, keepaspectratio]{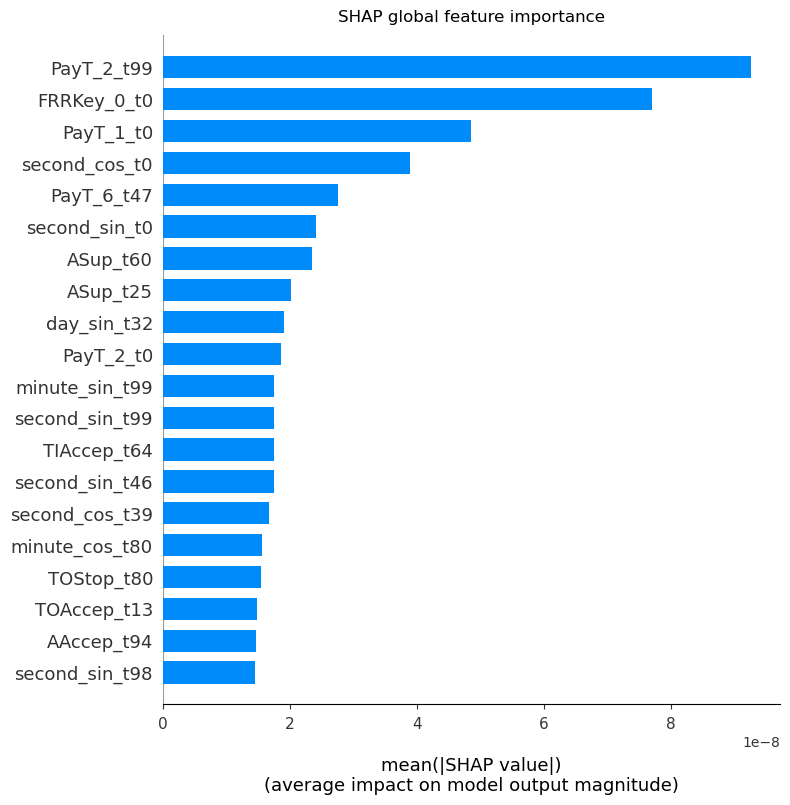}
  \caption{SHAP summary bar plot for original-label Transformer for the declaration-level dataset.}
  \label{fig:shaporigtransbased}
\end{minipage}
\hfill
\begin{minipage}[t]{0.48\textwidth}
  \centering
  \includegraphics[width=0.8\linewidth, height=8cm, keepaspectratio]{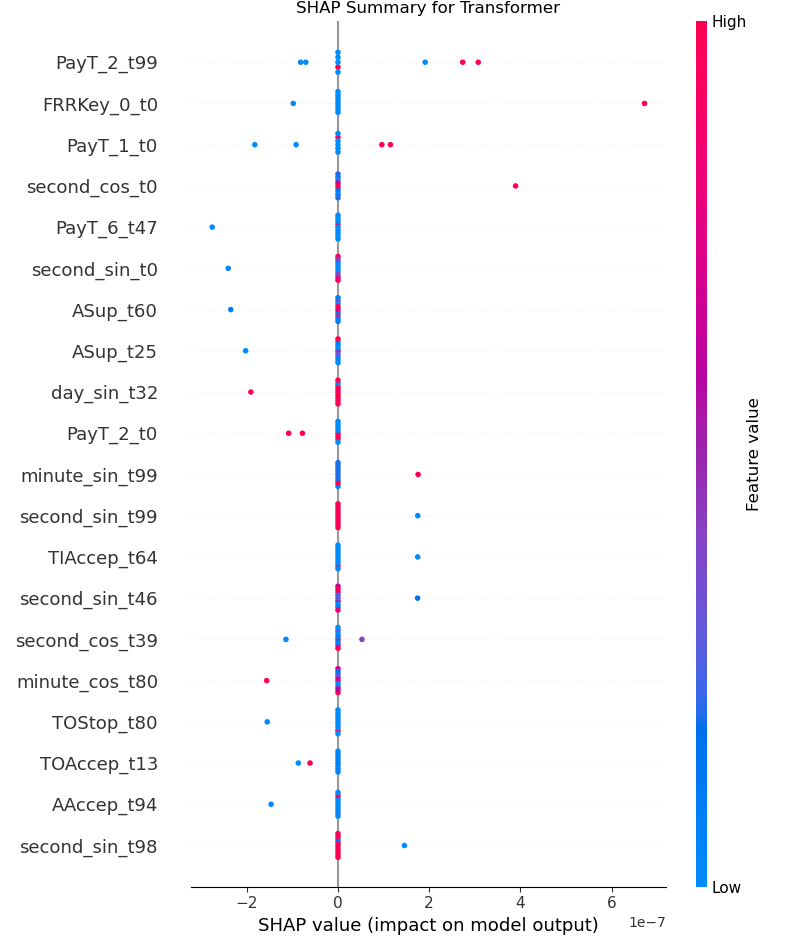}
  \caption{SHAP summary plot for original-label Transformer for the declaration-level dataset.}
  \label{fig:shapsumorigtransbased}
\end{minipage}
\end{figure}

\begin{figure}[h!]
\centering
\begin{minipage}[t]{0.48\textwidth}
  \centering
  \includegraphics[width=0.8\linewidth, height=8cm, keepaspectratio]{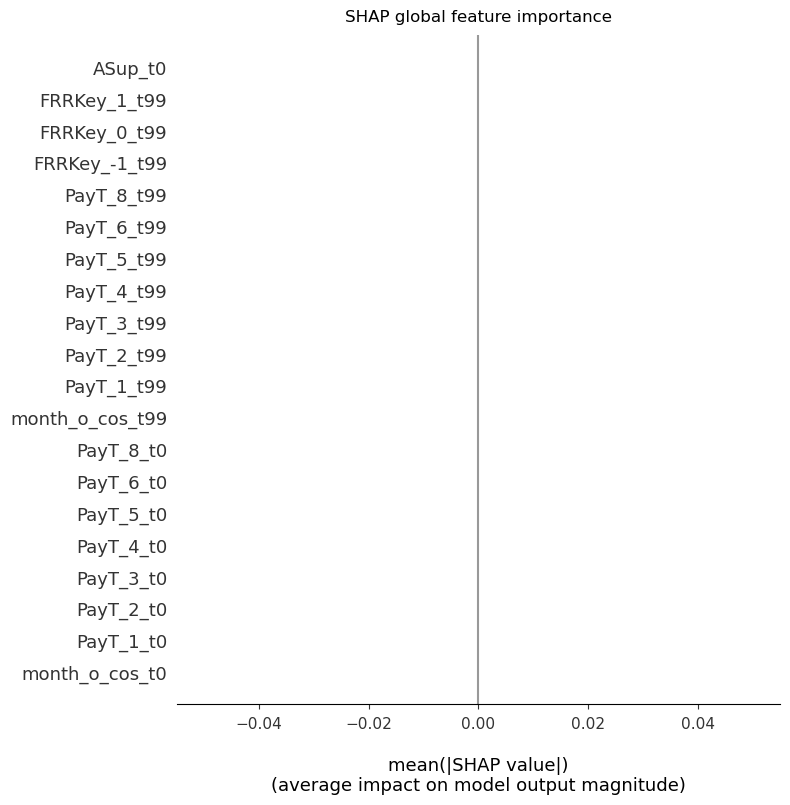}
  \caption{SHAP summary bar plot for original-label Transformer for the operation-level dataset.}
  \label{fig:shaporigtransbasedo}
\end{minipage}
\hfill
\begin{minipage}[t]{0.48\textwidth}
  \centering
  \includegraphics[width=0.8\linewidth, height=8cm, keepaspectratio]{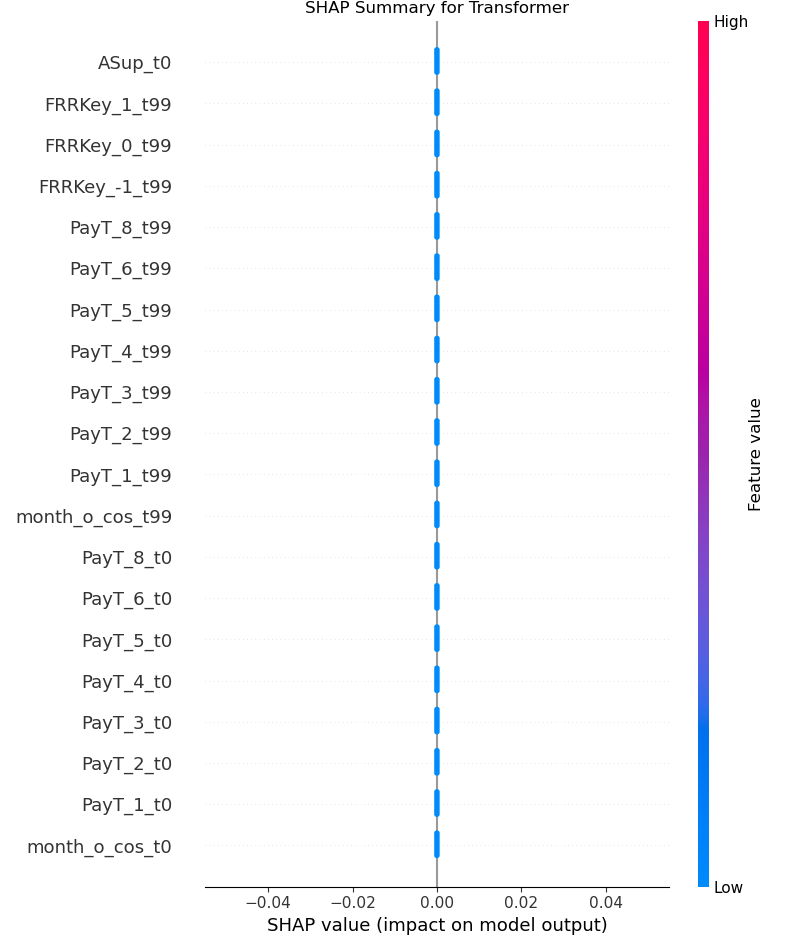}
  \caption{SHAP summary plot for original-label Transformer for the operation-level dataset.}
  \label{fig:shapsumorigtransbasedo}
\end{minipage}
\end{figure}

\end{appendices}

\end{document}